\documentclass{bmvc2k}

\usepackage{times}
\usepackage{epsfig}
\usepackage{graphicx}
\usepackage{amsmath}
\usepackage{amssymb}
\usepackage{wrapfig}

\title{Image Composition Assessment with Saliency-augmented Multi-pattern Pooling}

\addauthor{Bo Zhang}{bo-zhang@sjtu.edu.cn}{1}
\addauthor{Li Niu$^\ast$}{ustcnewly@sjtu.edu.cn}{1}
\addauthor{Liqing Zhang}{zhang-lq@cs.sjtu.edu.cn}{1}

\addinstitution{
 MoE Key Lab of Artificial Intelligence \\
 Shanghai Jiao Tong University \\
 Shanghai, China
}

\runninghead{Zhang, Niu, Zhang}{Image Composition Assessment with SAMP}

\begin{document}

\maketitle
\begin{abstract}
Image composition assessment is crucial in aesthetic assessment, which aims to assess the overall composition quality of a given image. However, to the best of our knowledge, there is neither dataset nor method specifically proposed for this task. In this paper, we contribute the first composition assessment dataset CADB with composition scores for each image provided by multiple professional raters. Besides, we propose a composition assessment network SAMP-Net with a novel Saliency-Augmented Multi-pattern Pooling (SAMP) module, which analyses visual layout from the perspectives of multiple composition patterns. We also leverage composition-relevant attributes to further boost the performance, and extend Earth Mover's Distance (EMD) loss to weighted EMD loss to eliminate the content bias. The experimental results show that our SAMP-Net can perform more favorably than previous aesthetic assessment approaches.
\end{abstract}

\section{Introduction}
\label{sec:intro}
Image aesthetic assessment aims to judge aesthetic quality automatically in a qualitative or quantitative way, which can be widely used in many down-stream applications such as assisted photo editing, intelligent photo album management, image cropping, and smartphone photography \cite{Bhattacharya2011AHA,Chen2017LearningTC,Datta2006StudyingAI,rawat2015context,fang2020perceptual,tu2020image,rawat2017spring,rawat2016clicksmart}. 
Among the factors related to image aesthetics, image composition, which mainly concerns the arrangement of the visual elements inside the frame \cite{Prakel2010TheFO}, is very critical in estimating image aesthetics \cite{savakis2000evaluation,obrador2010role,Liu2020CompositionAwareIA}, because composition directs the attention of viewer and has a significant impact on the aesthetic perception \cite{Prakel2010TheFO,Freeman2007ThePE,Martnez1988VisualFA}. 

Despite the importance of image composition, there is no dataset readily available for image composition assessment. Some existing aesthetic datasets contain annotations related to image composition \cite{Kong2016PhotoAR, Murray2012AVAAL, Jin2019AestheticAA, Chang2017AestheticCG}.
\textcolor[rgb]{0,0,0}{However, they only have composition-relevant attributes without overall composition score except for PCCD dataset \cite{Chang2017AestheticCG}, but PCCD dataset only presents one reviewer's composition rating for each image and this reviewer, an anonymous website visitor, may be unprofessional. So the ratings might be biased and inaccurate, which are far below the requirement for scientific evaluation.}
To this end, we contribute a new image Composition Assessment DataBase (CADB) on the basis of Aesthetics and Attributes DataBase (AADB) dataset \cite{Kong2016PhotoAR}. Our CADB dataset contains 9,497 images with each image rated by 5 individual raters who specialize in fine art for the overall composition quality. The details of our CADB dataset will be introduced in Section~\ref{sec:dataset}.


\begin{wrapfigure}{R}{0.5\linewidth}
  \vspace{-7mm}
  \begin{center}
    \includegraphics[width=1\linewidth]{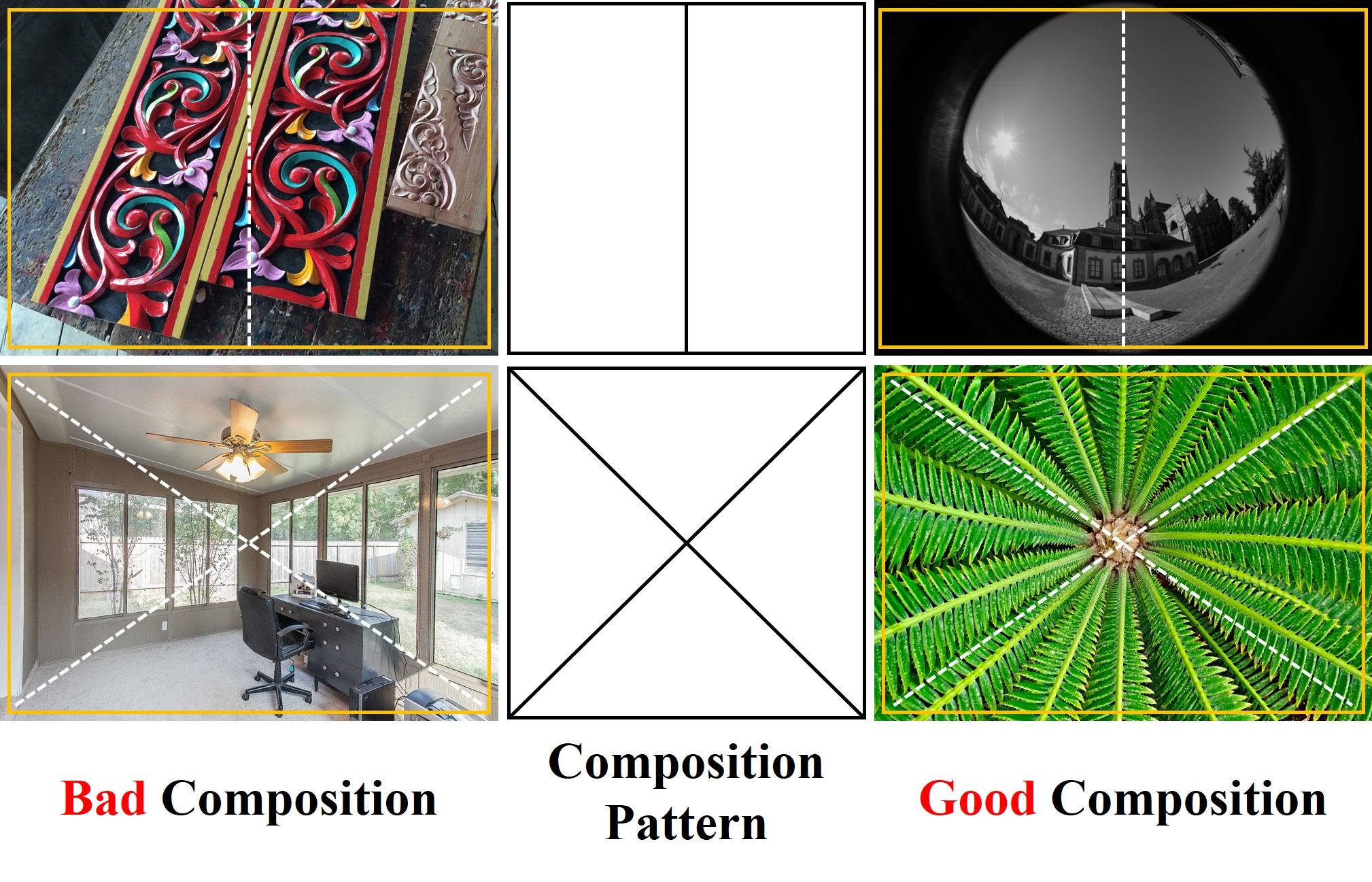}
  \end{center}
  \vspace{-6mm}
  \caption{Evaluating composition quality from the perspectives of different composition patterns. The first (\emph{resp.}, second) row shows a good example and a bad example considering symmetrical (\emph{resp.}, radial) balance.}
  \vspace{-2mm}
  \label{fig:pattern_example}
\end{wrapfigure}

To the best of our knowledge, there is no method specifically designed for image composition assessment. However, some previous aesthetic assessment methods also take composition into consideration. We divide the existing composition-relevant approaches into two groups. 1) The composition-preserving methods \cite{Mai2016CompositionPreservingDP, Chen2020AdaptiveFD} can maintain image composition during both training and testing. However, these approaches fail to extract composition-relevant feature for composition assessment task. 2) The composition-aware approaches \cite{Liu2020CompositionAwareIA, Ma2017ALampAL, Wang2019ModelingHP} extract composition-relevant feature by modeling the mutual dependencies between all pairs of objects or regions in the image. However, redundant and noisy information is likely to be introduced during this procedure, which may adversely affect the performance of composition assessment. Moreover, there are some previous methods \cite{wu2017high,Dhar2011HighLD,Bhattacharya2010AFF,wu2010good,Tang2013ContentBasedPQ,liu2010optimizing} designed to model the well-established photographic rules (\emph{e.g.}, rule of thirds and golden ratio \cite{joshi2011aesthetics}), which humans use in evaluating image composition quality. 
\textcolor[rgb]{0,0,0}{However, these rule-based methods have two major limitations: 1) The hand-crafted feature extraction is tedious and laborious compared with deep learning features \cite{li2020novel}. 2) Each rule is valid only for specific scenes and they did not consider which rules are applicable for a given scene \cite{su2021camera}. }

Interestingly, composition pattern, as an important aspect of composition assessment, is not explicitly considered by the above methods. As shown in Figure~\ref{fig:pattern_example}, each composition pattern divides the holistic image into multiple non-overlapping partitions, which can model human perception of composition quality. In particular, by analyzing the visual layout (\emph{e.g.}, positions and sizes of visual elements) according to composition pattern, \emph{i.e.}, comparing the visual elements in various partitions, we can quantify the aesthetics of visual layout in terms of visual balance (\emph{e.g.}, symmetrical balance and radial balance) \cite{jahanian2015learning, lok2004evaluation,Lee2017SemanticLD}, composition rules (\emph{e.g.}, rule of thirds, diagonals and triangles) \cite{thommes2018instagram,Lee2018PhotographicCC}, and so on. Different composition patterns offer different perspectives to evaluate composition quality. 
For example, the composition pattern in the top (\emph{resp.}, bottom) row in Figure~\ref{fig:pattern_example} can help judge the composition quality in terms of symmetrical (\emph{resp.}, radial) balance.

To dissect visual layout based on different composition patterns, we propose a novel multi-pattern pooling module at the end of backbone to integrate the information extracted from multiple patterns, in which each pattern provides a perspective to evaluate the composition quality. Considering that the sizes and locations of salient objects are representative of visual layout and fundamental to image composition \cite{lok2004evaluation}, we further integrate visual saliency \cite{Hou2007SaliencyDA} into our multi-pattern pooling module to encode the spatial and geometric information of salient objects, leading to our Saliency-Augmented Multi-pattern Pooling (SAMP) module. Additionally, since some composition patterns may play more important roles, we design weighted multi-pattern aggregation to fuse multi-pattern features, which can adaptively assign different weights to different patterns.  

Moreover, because our dataset is built upon AADB dataset \cite{Kong2016PhotoAR} with composition-relevant attributes, we further leverage composition-relevant attributes to boost the performance of composition assessment. Specifically, we propose an Attentional Attribute Feature Fusion (AAFF) module to fuse composition feature and attribute feature. Finally, after noticing the content bias existing in our dataset, that is, composition score distribution is severely influenced by object category, we extend Earth Mover's Distance (EMD) loss in \cite{Hou2016SquaredEM} to weighted EMD loss to eliminate the content bias. 

The main contributions of this paper can be summarized as follows: 1) We contribute the first image composition assessment dataset CADB, in which each image has the composition scores annotated by five professional raters. 2) We propose a novel composition assessment method with Saliency-Augmented Multi-pattern Pooling (SAMP) module. 3) We investigate the effectiveness of auxiliary attributes and weighted EMD loss for composition assessment. 4) Our model outperforms previous aesthetic assessment methods on our dataset. 
\vspace{-3mm}
\section{Related Work}

\subsection{Aesthetic Assessment Dataset}
Many large-scale aesthetic assessment datasets have been collected in recent years, like Aesthetic Visual Analysis database (AVA) \cite{Murray2012AVAAL}, AADB \cite{Kong2016PhotoAR}, Photo Critique Captioning Dataset (PCCD) \cite{Chang2017AestheticCG}, AVA-Comments \cite{Zhou2016JointIA}, AVA-Reviews \cite{Wang2019NeuralAI}, FLICKER-AES \cite{Ren2017PersonalizedIA}, and DPC-Captions \cite{Jin2019AestheticAA}.
However, these datasets only have composition-relevant attributes without overall composition score, or only have one inaccurate composition score for each image, which are far below the requirement for composition assessment research.
Unlike the existing aesthetic datasets, our CADB dataset contains composition ratings assigned to each image by multiple professional raters. Besides, we guarantee the reliability of our dataset based on sanity check and consistency analysis (see Section~\ref{sec:dataset}).

\begin{figure*}[tbp]
\begin{center}
   \includegraphics[width=1\linewidth]{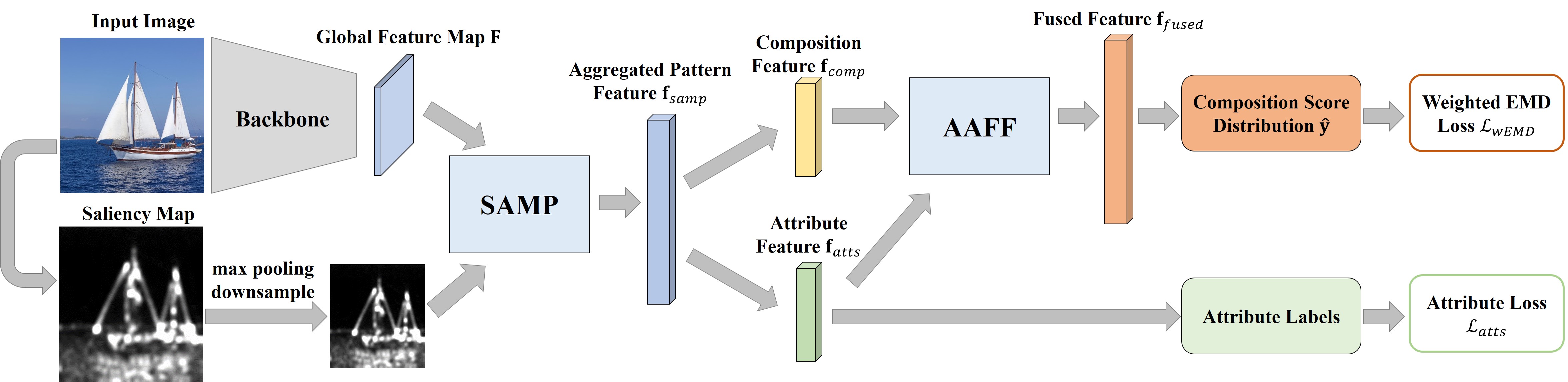}
\end{center}
   \vspace{-5mm}
   \caption{The overall pipeline of our SAMP-Net for composition assessment. We use ResNet18 \cite{he2016deep} as backbone. The detailed structure of our Saliency-Augmented Multi-pattern Pooling (SAMP) module and Attentional Attribute Feature Fusion (AAFF) module are illustrated in Figure~\ref{fig:pattern_and_samp} and Figure~\ref{fig:aaff} respectively.}
  \vspace{-5mm}
\label{fig:pipeline}
\end{figure*}
\subsection{Composition-relevant Aesthetic Assessment}
 We can divide existing composition-relevant aesthetic assessment methods into traditional methods and deep learning methods. 
 As surveyed in \cite{Deng2017ImageAA, brachmann2017computational}, traditional methods \cite{liu2010optimizing,Tang2013ContentBasedPQ, Bhattacharya2010AFF, Zhang2014FusionOM, Murray2012AVAAL, Zhou2015ModelingPE,Su2011ScenicPQ, Li2010AestheticQA, wu2010good,Marchesotti2011AssessingTA,obrador2010role,savakis2000evaluation,rawat2015context} usually employed hand-crafted features or generic image features (\emph{e.g.}, bag-of-visual-words \cite{Su2011ScenicPQ} and Fisher vectors \cite{Perronnin2007FisherKO}) to learn image aesthetic evaluation, yet their generalization ability is limited by the complexity of image composition assessment task.
 The deep learning based methods can be divided into two groups. The composition-preserving approaches \cite{Mai2016CompositionPreservingDP, Chen2020AdaptiveFD}, without explicitly learning composition representations, produce inferior results on composition evaluation task.
 The composition-aware approaches \cite{Liu2020CompositionAwareIA, Ma2017ALampAL, Wang2019ModelingHP} consider the relationship between all pairs of objects or regions in the image for modeling image composition, which is likely to introduce redundant and noisy information.
 Moreover, the above methods did not explicitly consider composition patterns. In contrast, we design a novel Saliency-Augmented Multi-pattern Pooling (SAMP) module, which provides an insightful and effective perspective for evaluating composition quality.

\begin{figure}[tbp]
\begin{center}
  \includegraphics[width=0.9\linewidth]{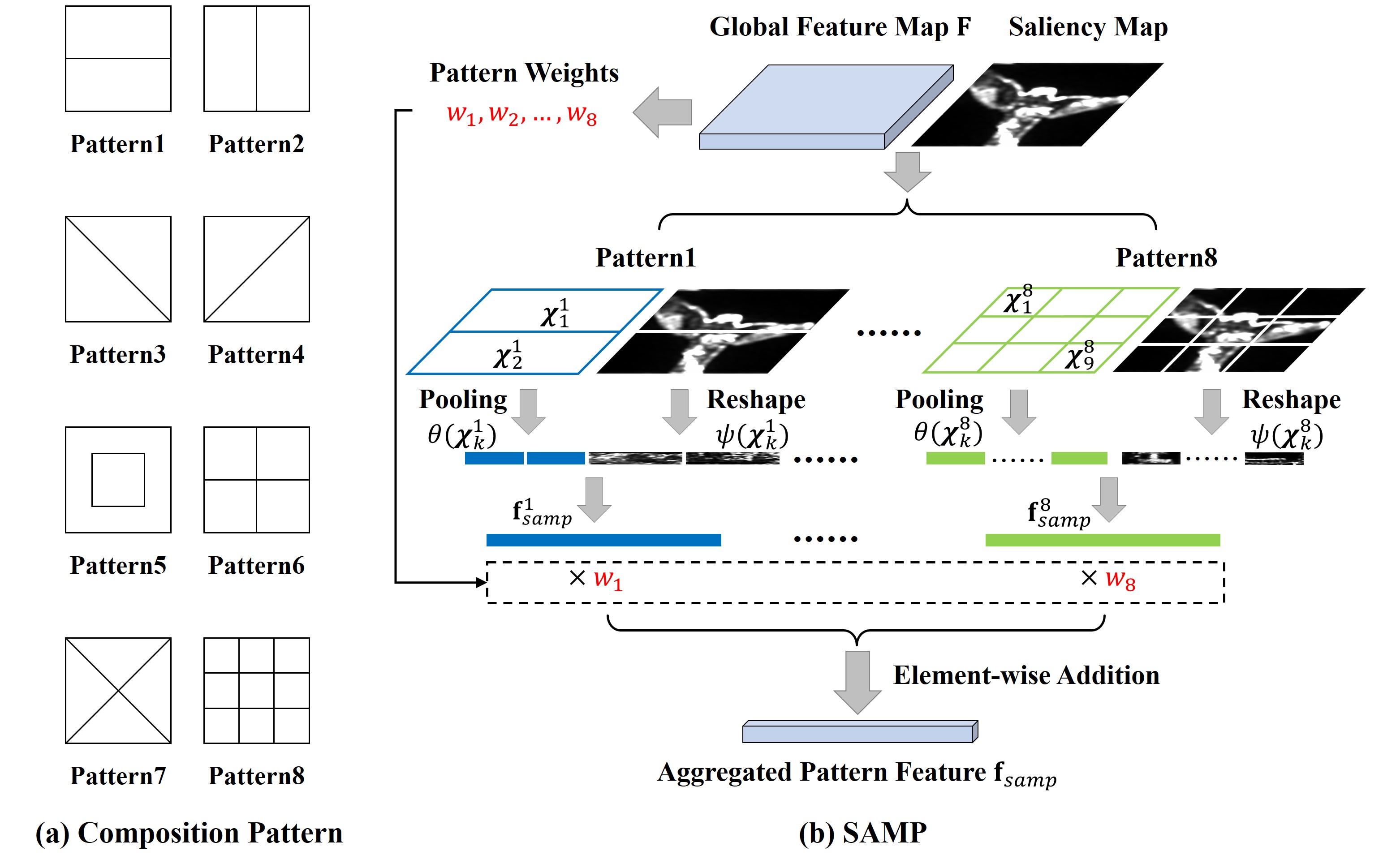}
\end{center}
  \vspace{-6mm}
  \caption{Our designed eight composition patterns and Saliency-augmented Multi-pattern Pooling (SAMP) module.}
  \vspace{-4mm}
\label{fig:pattern_and_samp}
\end{figure}
\section{Composition Assessment DataBase (CADB)}
\label{sec:dataset}
To the best of our knowledge, there is no prior dataset specifically constructed for composition assessment. To support the research on this task, we build a dataset upon the existing AADB dataset \cite{Kong2016PhotoAR}, from which we collect a total of 9,958 real-world photos. We adopt a composition rating scale from 1 to 5, where a larger score indicates better composition.  We make annotation guidelines for composition quality rating and train five individual raters who specialize in fine art.  So for each image, we can obtain five composition scores ranging from 1 to 5.
Given the subjective nature of human aesthetic activity \cite{savakis2000evaluation,Prakel2010TheFO,Freeman2007ThePE}, we perform sanity check and consistency analysis. Similar to \cite{Yu2018LearningTD}, we use 240 additional ``sanity check'' images during annotating to roughly verify the validness of our annotations. We also examine the consistency of composition ratings provided by five individual raters  (see Supplementary).  Similar to \cite{Murray2012AVAAL, Kong2016PhotoAR}, we average the composition scores as the ground-truth composition mean score for each image, which is denoted as $\bar{y}$. More details about our CADB dataset will be elaborated in Supplementary. 

 Besides, we observe the content bias in our CADB dataset, that is, there are some biased categories whose score distributions are concentrated in a very narrow interval. After removing 461 biased images, we split the remaining images into 8,547 training images and 950 test images, in which the test set is made less biased for better evaluation (see Supplementary).

\section{Methodology}
\label{sec:methodology}
To accomplish the composition assessment task, we propose a novel network SAMP-Net, which is named after Saliency-Augmented Multi-pattern Pooling (SAMP) module. The overall pipeline of our method is illustrated in Figure~\ref{fig:pipeline}, where we first extract the global feature map from input image by backbone (\emph{e.g.}, ResNet18 \cite{he2016deep}) and then yield aggregated pattern feature through our SAMP module, which is followed by Attentional Attribute Feature Fusion (AAFF) module to fuse the composition feature and attribute feature. After that, we predict composition score distribution based on the fused feature and predict the attribute score based on the attribute feature, which are supervised by weighted EMD loss and Mean Squared Error (MSE) loss respectively.
\subsection{Saliency-augmented Multi-pattern Pooling}

\noindent\textbf{Multi-pattern Pooling:} As demonstrated in Figure~\ref{fig:pattern_and_samp}(a), we empirically design eight basic composition patterns inspired by classic composition guidelines. \textcolor[rgb]{0,0,0}{For instance, pattern 1,2,6,7 are inspired by symmetrical composition.
Pattern 3,4 are inspired by diagonal composition.
Pattern 5 is inspired by centre composition.
Pattern 8 is inspired by rule of thirds \cite{thommes2018instagram,Lee2018PhotographicCC}.
Although our pattern design is inspired by composition rules, there is no strict one-to-one correspondence between composition rules and patterns. Each pattern provides a perspective for evaluating composition quality, which may be beyond the scope of a single rule.  For example, Pattern 8 is related to rule of thirds, but not limited to rule of thirds. Based on Pattern 8, more useful information can be excavated by comparing the visual elements in nine partitions.}

Since humans typically employ multiple perspectives when analysing image composition, composition assessment should be accomplished based on all composition patterns in a comprehensive way.
Therefore, we propose multi-pattern pooling to achieve this goal, which is illustrated in Figure~\ref{fig:pattern_and_samp}(b). Given an $H \times W$ global feature map $\mathbf{F}$ with $C$ channels, which is extracted from input image by backbone, we represent the pixel-wise feature at each location as $\mathbf{x}_{i,j}$, where $0 < i \leq H$, $0 < j \leq W$. For the $p$-th pattern, we divide $\mathbf{F}$ into $K_p$ non-overlapping partitions $\{\mathcal{X}^p_1, \mathcal{X}^p_2, \ldots, \mathcal{X}^p_{K_p} \}$ and $K_p$ is the total number of partitions in this pattern. Then, the feature of the $k$-th partition can be obtained via average pooling: $\theta(\mathcal{X}^p_k) = \frac{1}{| \mathcal{X}^p_k |} \sum_{ (i,j) \in \mathcal{X}^p_k} \mathbf{x}_{i,j}\in \mathbb{R}^{C}$.


\noindent\textbf{Saliency-augmented Multi-pattern Pooling:} 
Considering the significance of salient objects for composition assessment, we further incorporate the saliency information (\emph{i.e.}, locations and scales of salient objects) into multi-pattern pooling. 
To achieve this goal, we utilize an unsupervised saliency detection method \cite{Hou2007SaliencyDA} to produce saliency maps for input images. We have also tried several supervised methods \cite{hou2017deeply,cornia2018predicting,Zhao_2019_CVPR}, which prove to be less effective. 
After obtaining the saliency map, we downsample it to $H_{sal} \times W_{sal}$ through max pooling. Recall that the size of global feature map is $H \times W$, we set $H_{sal} = 8H$ and $W_{sal} = 8W$ for retaining more details of salient objects. 

Different from $\theta(\mathcal{X}^p_k)$ using average pooling,  we directly reshape each partition of saliency map into a vector, because the pooling operation will result in significant information loss.
Specifically, for the $k$-th partition in the $p$-th pattern, we reshape the saliency map in this partition into a saliency vector $\psi(\mathcal{X}_k^p) \in \mathbb{R}^{D^{p}_k}$, in which $D^{p}_k$ varies with partition and pattern. Then, we concatenate $\psi(\mathcal{X}_k^p)$ and $\theta(\mathcal{X}^p_k)$ to generate the partition feature $[\psi{(\mathcal{X}^p_k)}, \theta(\mathcal{X}^p_k)]$. 

For the $p$-th pattern, we concatenate the partition features of $K_p$ partitions into a long vector $\tilde{\mathbf{f}}^p_{samp}$, which is followed by a fc layer and $\mathrm{ReLU}$ activation function to produce the pattern vector $\mathbf{f}^p_{samp} \in \mathbb{R}^{C^{\prime}}$.  \textcolor[rgb]{0,0,0}{Intuitively, $[\psi(\mathcal{X}_k^p),\theta(\mathcal{X}^p_k)]$ extracts the visual information in each partition and $\mathbf{f}^p_{samp}$  encodes the relationship among visual elements in different partitions.}

\noindent\textbf{Weighted Multi-pattern Aggregation:} 
Since some composition patterns may play more important roles when evaluating image composition, our model is trained to assign different weights for different patterns. Precisely, we apply global average pooling, a fc layer, and $\mathrm{softmax}$ normalization to the global feature map $\mathbf{F}$, producing the multi-pattern weight $w_p$ for the $p$-th pattern. Then, we have the aggregated pattern feature via weighted summation $\mathbf{f}_{samp} = \sum^P_{p=1} w_p \mathbf{f}^p_{samp}$, in which $P$ is the number of composition patterns ($P=8$). Based on the learnt weights, we can know the dominant patterns in determining the overall composition quality and provide interpretable guidance for users (see Section~\ref{pattern_analysis}).

\noindent\textbf{Comparison with Spatial Pyramid Pooling}: Although the proposed SAMP and Spatial Pyramid Pooling (SPP) \cite{He2015SpatialPP} are similar in architecture, both of which pool features from multiple sets of partitions, SAMP is significantly different from SPP in three main aspects: 1) our pooling patterns are specifically designed and well-tailored for image composition evaluation, which can analyse the composition quality from the viewpoint of composition patterns; 2) we introduce visual saliency into multi-pattern pooling; 3) we learn pattern weights which provide interpretable guidance for improving composition quality.

\begin{wrapfigure}{R}{0.6\linewidth}
  \vspace{-8mm}
  \begin{center}
    \includegraphics[width=1\linewidth]{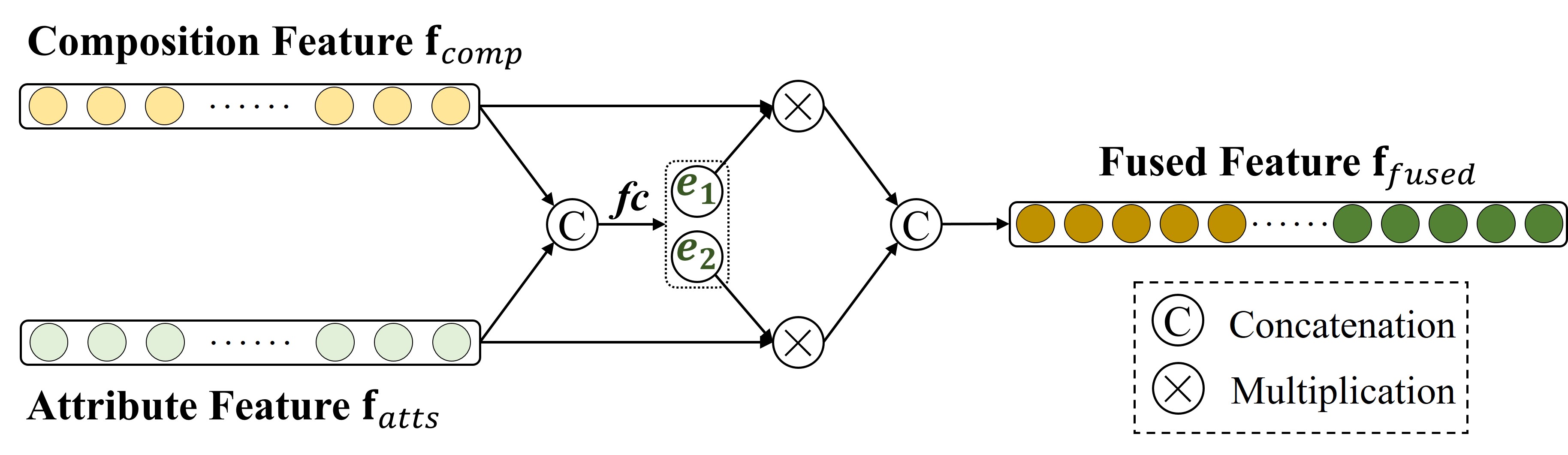}
  \end{center}
  \vspace{-5mm}
  \caption{Attentional Attribute Feature Fusion (AAFF) module. \emph{fc} means a fully-connected layer with $\mathrm{sigmoid}$ activation and $e_1, e_2$ are attention coefficients.}
  \label{fig:aaff}
\end{wrapfigure}
\subsection{Attentional Attribute Feature Fusion}
\label{aaff}
Since our dataset is built upon AADB \cite{Kong2016PhotoAR}, which is associated with composition-relevant attributes, it is natural to consider using them to help composition assessment. We use five composition-relevant attribute annotations: \emph{rule of thirds}, \emph{balancing elements}, \emph{object emphasis}, \emph{symmetry}, and \emph{repetition}. 

Specifically, as illustrated in Figure~\ref{fig:pipeline}, we decompose the aggregated pattern feature $\mathbf{f}_{samp} \in \mathbb{R}^{C^{\prime}}$ into composition feature $\mathbf{f}_{comp}$ and attribute feature $\mathbf{f}_{atts}$ by using two separate fc layers, the dimensions of which are both set to $\frac{C^{\prime}} {2}$. 
We dynamically weigh the contributions of $\mathbf{f}_{comp}$ and $\mathbf{f}_{atts}$ for the composition assessment task, as illustrated in Figure~\ref{fig:aaff}. First, we apply a fc layer and $\mathrm{sigmoid}$ activation to the concatenation of  $\mathbf{f}_{comp}$ and $\mathbf{f}_{atts}$, to learn the attention coefficients $[e_1, e_2]$ for two types of features. Then, we concatenate the weighted composition feature and attribute feature, yielding the fused feature $\mathbf{f}_{fused} = [e_1 \mathbf{f}_{comp}, e_2 \mathbf{f}_{atts} ] \in \mathbb{R}^{C^{\prime}}$. During training, an additional layer is added to perform attribute prediction based on the attribute feature $\mathbf{f}_{atts}$. We employ MSE loss denoted as $\mathcal{L}_{atts}$ for attribute prediction.

As mentioned in Section~\ref{sec:dataset}, we observe the content bias in our dataset, in which case the network may find a shortcut to simply rate images based on their contents.
To mitigate the content bias in training set, we extend EMD loss to weighted EMD loss denoted as  $\mathcal{L}_{wEMD}$ (see Supplementary), which assigns smaller weights to biased samples when calculating EMD Loss.
Finally, our SAMP-Net can be trained in an end-to-end manner with attribute prediction loss $\mathcal{L}_{atts}$ and weighted EMD loss $\mathcal{L}_{wEMD}$:
\begin{equation}
    \mathcal{L} = \mathcal{L}_{wEMD} + \lambda \mathcal{L}_{atts},
    \label{total_loss}
\end{equation}
where $\lambda$ is a trade-off parameter set as $0.1$ via cross validation.

\vspace{-3mm}
\section{Experiments}

\begin{table*}
    \begin{center}
    \setlength{\tabcolsep}{2.5mm}{
        \begin{tabular}{|l c c c c c c|c|c|c|c|}
            \hline
            &  \emph{WE}&        \emph{MP}&   \emph{PW}&  \emph{SA}&  \emph{AF}& \emph{AA}&    MSE$\downarrow$   &EMD$\downarrow$    &SRCC$\uparrow$     &LCC$\uparrow$ \\
            \hline\hline            
            1&      &               &           &           &           &           &           0.4534           & 0.1943            & 0.6025            & 0.6148       \\
            2&      \checkmark&     &           &           &           &           &           0.4373           & 0.1859            & 0.6105            & 0.6258       \\
            \hline
            3&      \checkmark& \checkmark&     &           &           &           &           0.4170           & 0.1847            & 0.6292            & 0.6435       \\
            4&      \checkmark& \checkmark& \checkmark&     &           &           &           0.4134           & 0.1829            & 0.6323            & 0.6483       \\
            5&      \checkmark& \checkmark& \checkmark& \checkmark&     &           &           0.4088           & 0.1820            & 0.6421            & 0.6544       \\
            6&      \checkmark& $\dagger$&  \checkmark& \checkmark&     &           &           0.4274           & 0.1854            & 0.6226            & 0.6293       \\
            7&      \checkmark& $\ddagger$& \checkmark& \checkmark&     &           &           0.4205           & 0.1845            & 0.6319            & 0.6363       \\
            \hline
            8&      \checkmark&     &           &           &       \checkmark&     &           0.4320           & 0.1850            & 0.6200            & 0.6303       \\
            9&      \checkmark& \checkmark& \checkmark& \checkmark& \checkmark&     &           0.3979           & 0.1817            & 0.6439            & 0.6610       \\
            10&     \checkmark& \checkmark& \checkmark& \checkmark& \checkmark& \checkmark& \textbf{0.3867}    &\textbf{0.1798}    &\textbf{0.6564}    &\textbf{0.6709} \\
            \hline
        \end{tabular}}
    \end{center}
    \caption{Ablation studies of different components in our model. $\dagger$ means Spatial Pyramid Pooling (SPP) \cite{He2015SpatialPP}. $\ddagger$ means Multi-scale Pyramid Pooling (MPP) \cite{yoo2015multi}. \emph{WE} means weighted EMD loss. \emph{MP} means multi-pattern pooling. \emph{PW} means pattern weights. \emph{SA} means saliency-augmented. \emph{AF} indicates attribute feature and \emph{AA} indicates attentional attribute feature fusion.}
    \label{table_ablation}
    \vspace{-5mm}
\end{table*}

\subsection{Implementation Details and Evaluation Metric}
\label{sec:implement_and_metric}
We use ResNet18 \cite{he2016deep} pretrained on ImageNet \cite{deng2009imagenet} as the backbone of our SAMP-Net. Unless otherwise specified, all input images are resized to $224 \times 224$ for both training and testing following \cite{li2020personality,schwarz2018will,ko2018pac}, leading to a global feature map of $H\times W= 7 \times 7$, and the saliency map is downsampled to $H_{sal} \times W_{sal}= 56 \times 56$ before passing to the SAMP. More details can be found in Supplementary. All experiments are conducted on our CADB dataset. 

To evaluate the composition score distribution and composition mean score predicted by different models, it is natural to adopt EMD and MSE as the evaluation metrics. EMD measures the closeness between the predicted and ground-truth composition score distributions as in \cite{Hou2016SquaredEM}. MSE is computed between the predicted and ground-truth  composition mean scores. Moreover, following existing aesthetic assessment approaches \cite{Kong2016PhotoAR, Talebi2018NIMANI, Chen2020AdaptiveFD}, we also report the ranking correlation measured by Spearman's Rank Correlation Coefficient (SRCC) and the linear association measured by Linear Correlation Coefficient (LCC) between the predicted and ground-truth composition mean scores.  
\vspace{-2mm}
\subsection{Ablation Study}
\label{sec:ablation}

To evaluate the effectiveness of each individual component in our SAMP-Net, we conduct a series of experiments and report all the evaluation metrics described in Section~\ref{sec:implement_and_metric}. In this section, we start from ResNet18 backbone and build up our holistic model step by step.

\noindent\textbf{Weighted EMD Loss:} We start from basic ResNet18 \cite{he2016deep}, and report the results using EMD loss and weighted EMD loss in Table~\ref{table_ablation}. The experimental results show that training with weighted EMD loss (row 2) performs better than standard EMD loss (row 1) with a clear gap of test EMD between these two models, which is attributed to the advantage of weighted EMD loss in eliminating content bias. 

\noindent\textbf{Saliency-Augmented Multi-pattern Pooling (SAMP):} Based on ResNet18 with weighted EMD loss (row 2), we add our SAMP module and also explore its ablated versions. 
We first investigate vanilla multi-pattern pooling without saliency or pattern weights (row 3), in which saliency vector is excluded from partition feature and the pattern features of multiple patterns are simply averaged. 
Then, we learn pattern weights to aggregate multiple pattern features (row 4). By comparing row 3 and row 4,  it is beneficial to adaptively assign different weights to different pattern features. 
We further incorporate saliency map into SAMP module (row 5). The comparison between row 4 and row 5 proves that is useful to emphasize the layout information of salient objects.
Considering the architecture similarity between Spatial Pyramid Pooling (SPP) \cite{He2015SpatialPP} and our multi-pattern pooling, we replace our multi-pattern pooling with SPP using scales $\{1 \times 1, 2 \times 2, 3 \times 3\}$ following \cite{Chen2020AdaptiveFD} (row 6).
\textcolor[rgb]{0,0,0}{In addition, we also show the results of using Multi-scale Pyramid Pooling (MPP) \cite{yoo2015multi} in row 7, in which we make an image pyramid containing three scaled images. The comparisons (row 5 \emph{v.s.} row 6, row 5 \emph{v.s.} row7) show that the model using multi-pattern pooling outperforms both SPP and MPP.}
The reason is that our proposed multi-pattern pooling is specifically designed and well-tailored for composition assessment task.

\noindent\textbf{Attentional Attribute Feature Fusion (AAFF):} Built on row 2 (\emph{resp.}, row 5) in Table~\ref{table_ablation}, we additionally learn attribute feature and directly concatenate it with composition feature, leading to  row 8 (\emph{resp.}, row 9). The experimental results demonstrate that composition-relevant attributes can help boost the performance of composition evaluation. This sheds light on that composition-relevant attribute prediction and composition evaluation are two related and reciprocal tasks. 
Finally, we complete our attentional attribute feature fusion module by learning weights for weighted concatenation (row 10). 
From row 9 and row 10, we can observe that the model using weighted concatenation is better than that using plain concatenation, which validates the superiority of attentional fusion mechanism.

\subsection{Comparison with Existing Methods}
\begin{table}
    \begin{center}
    \setlength{\tabcolsep}{5.5mm}{
        \begin{tabular}{|l|c|c|c|c|}
            \hline
            Method                                          & MSE$\downarrow$   &EMD$\downarrow$    &SRCC$\uparrow$     &LCC$\uparrow$ \\
            \hline\hline            
            ResNet18                                        & 0.4534            & 0.1943            & 0.6025            & 0.6148 \\
            \hline
            AADB \cite{Kong2016PhotoAR}                     & 0.4234            & 0.1923            & 0.6236            & 0.6415 \\
            MNA-CNN \cite{Mai2016CompositionPreservingDP}   & 0.4260            & 0.1944            & 0.6108            & 0.6375 \\
            A-Lamp \cite{Ma2017ALampAL}                     & 0.4230            & 0.1898            & 0.6270            & 0.6456 \\
            VP-Net \cite{Wang2019ModelingHP}                & 0.4304            & 0.1948            & 0.6169            & 0.6285 \\
            RG-Net \cite{Liu2020CompositionAwareIA}         & 0.4398            & 0.1915            & 0.6026            & 0.6218 \\
            AFDC-Net \cite{Chen2020AdaptiveFD}              & 0.4245            & 0.1910            & 0.6154            & 0.6388 \\
            \hline
            SAMP-Net (Ours)                                 &\textbf{0.3867}   &\textbf{0.1798}    &\textbf{0.6564}    &\textbf{0.6709} \\
            \hline
        \end{tabular}}
    \end{center}
    \caption{Comparison of different methods on the composition assessment task. All models are trained and evaluated on the proposed CADB dataset.}
    \label{table_baseline}
    \vspace{-4mm}
\end{table}
To the best of our knowledge, there is no method specifically designed for image composition assessment. Nevertheless, some previous aesthetic assessment methods \cite{Kong2016PhotoAR,Mai2016CompositionPreservingDP,Ma2017ALampAL,Wang2019ModelingHP, Liu2020CompositionAwareIA, Chen2020AdaptiveFD} explicitly take composition into consideration. Since most of these methods do not yield score distribution, we make a slight modification on the prediction layer of these methods to be compatible with EMD loss \cite{Hou2016SquaredEM}. For fair comparison, all methods are trained and tested on our CADB dataset with ResNet18 pretrained on ImageNet \cite{deng2009imagenet} as backbone.

In Table~\ref{table_baseline}, we compare our method with different composition-relevant aesthetic assessment methods.
The baseline model (ResNet18) only consists of the pretrained ResNet18 and a prediction head, which is the same as row 1 in Table~\ref{table_ablation}.
Among these baselines, A-Lamp is the most competitive one, probably because A-Lamp introduces additional saliency information to learn the pairwise spatial relationship between objects. 
Our SAMP-Net clearly outperforms all the composition-relevant baselines, which demonstrates that our method is more adept at image composition assessment.  

\subsection{Analysis of Composition Pattern}
\label{pattern_analysis}
\begin{figure}[tbp]
\begin{center}
  \includegraphics[width=1\linewidth]{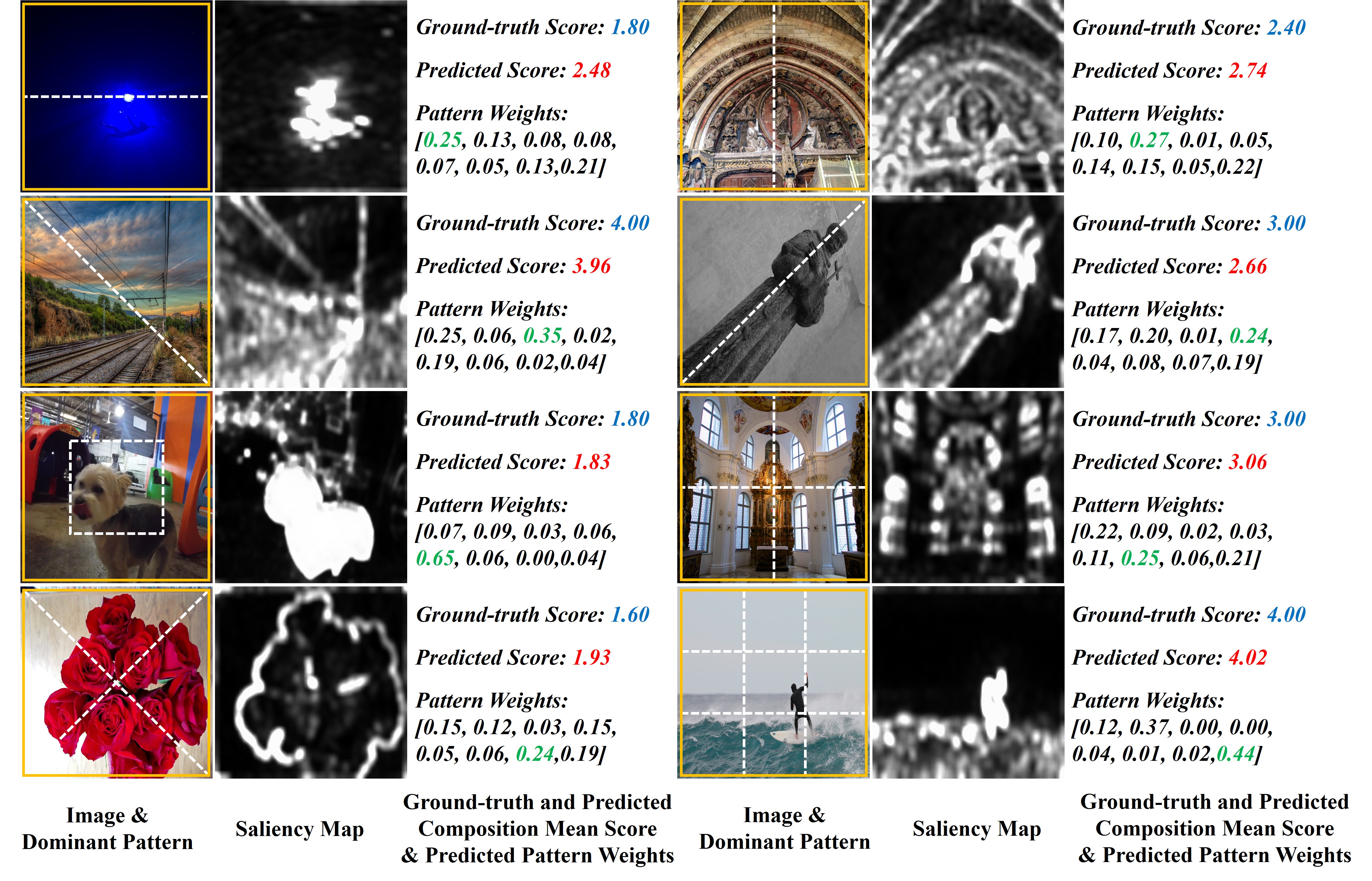}
\end{center}
  \vspace{-6mm}
   \caption{Analysis of the correlation between an image and its dominant pattern with the largest weight. We show the estimated pattern weights and the largest weight is colored green. We also show the ground-truth/predicted composition mean score in blue/red.}
  \vspace{-5mm}
\label{fig:model_interpretability}
\end{figure}


To take a close look at the learnt pattern weights which indicate the importance of different patterns on the overall composition quality, we show the input image, its saliency map, its ground-truth/predicted composition mean score, and its pattern weights in Figure~\ref{fig:model_interpretability}. 

For each image, the composition pattern with the largest weight is referred to as its dominant pattern. For each pattern, we show one example image with this pattern as dominant pattern and overlay this pattern on the image in Figure~\ref{fig:model_interpretability}, which reveals from which perspective the input image is given a high or low score.
For example, in the right figure of the last row, the surfer is placed at the intersection point between the gridlines of pattern 8, which implicates that the image conforms to the rule of thirds properly, yielding a relatively high score.
On the contrary, in the right figure of the first row, the arch slightly deviates from its symmetrical axis under pattern 2. So the low score implies that maintaining horizontal symmetry may enhance the composition quality.
In the left figure of the third row, per the low score under pattern 5, the dog is suggested to be moved to the center. 
In summary, our SAMP module can facilitate composition assessment by integrating the information from multiple patterns and provide constructive suggestions for improving the composition quality.
\vspace{-2mm}

\subsection{Additional Experiments in Supplementary}
Due to the space limitation, we present some experiments in Supplementary, including the results of using different training set sizes, backbones, and hyper-parameters $\lambda$ in  (\ref{total_loss}), weighted EMD loss analysis, the effectiveness of each pattern, the impact of using more composition patterns, comparison with the performance of human raters, more results on the CADB and PCCD \cite{Chang2017AestheticCG} datasets.
\vspace{-2mm}

\subsection{Limitations}
\label{sec:limitation}
\begin{figure}[tbp]
\begin{center}
  \includegraphics[width=1.0\linewidth]{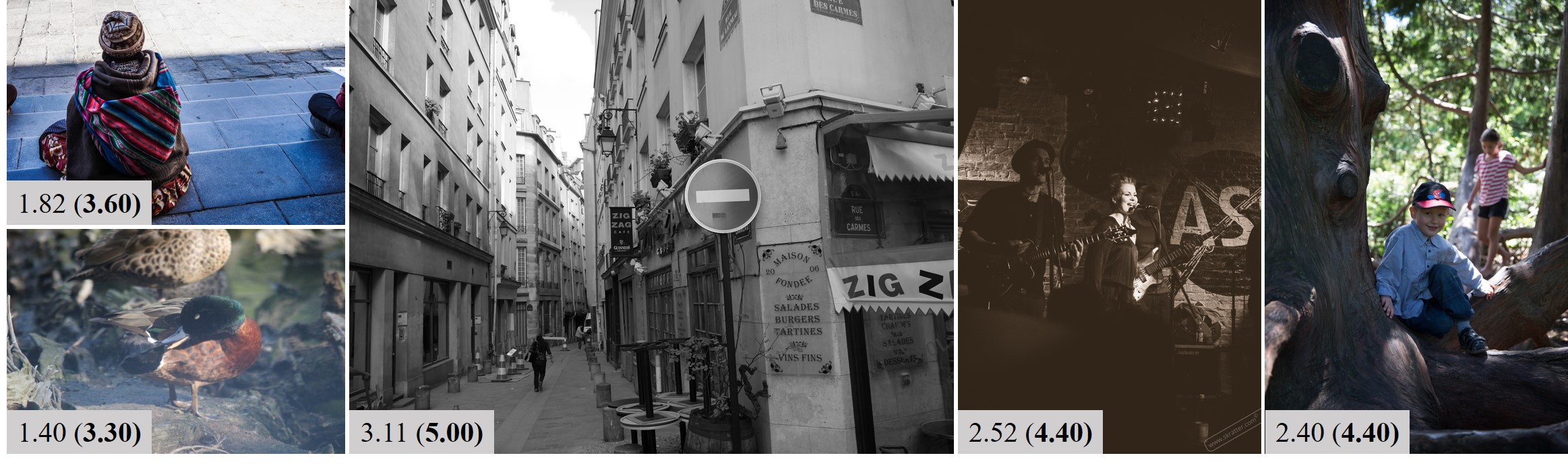}
\end{center}
  \vspace{-6mm}
   \caption{We show some failure cases in the test set,  which have the highest absolute errors between the predicted composition mean scores (out of bracket) and the ground-truth composition mean scores (in bracket). }
  \vspace{-5mm}
\label{fig:failure_cases}
\end{figure}

\textcolor[rgb]{0,0,0}{While our method can generally achieve accurate and reliable composition assessment, it still has some failure cases. We show several failure cases in Figure \ref{fig:failure_cases}, which have the highest absolute errors between the predicted and ground-truth composition mean scores. We can observe that our model tends to predict relatively low scores for these images with high composition mean scores, which is probably due to the distracting backgrounds and complicated composition patterns. 
In addition, there is a clear gap between our method and human raters on ranking the composition quality of different images (see Supplementary), which needs to be enhanced in the  future work.
}
\vspace{-4mm}

\section{Conclusion}
In this paper, we have contributed the first composition assessment dataset CADB with five composition scores for each image. 
We have also proposed a novel method SAMP-Net with saliency-augmented multi-pattern pooling. Equipped with SAMP module, AAFF module, and weighted EMD loss, our method is capable of achieving the best performance for composition assessment.
\vspace{-4mm}
\section*{Acknowledgement}
This work is  sponsored by National Natural Science Foundation of China (Grant No. 61902247) and Shanghai Sailing Program (19YF1424400).

\bibliography{egbib}

\begin{thebibliography}{61}
\providecommand{\natexlab}[1]{#1}
\providecommand{\url}[1]{\texttt{#1}}
\expandafter\ifx\csname urlstyle\endcsname\relax
  \providecommand{\doi}[1]{doi: #1}\else
  \providecommand{\doi}{doi: \begingroup \urlstyle{rm}\Url}\fi

\bibitem[Bhattacharya et~al.(2010)Bhattacharya, Sukthankar, and
  Shah]{Bhattacharya2010AFF}
S.~Bhattacharya, R.~Sukthankar, and M.~Shah.
\newblock A framework for photo-quality assessment and enhancement based on
  visual aesthetics.
\newblock In \emph{ACM-Multimedia}, 2010.

\bibitem[Brachmann and Redies(2017)]{brachmann2017computational}
A.~Brachmann and C.~Redies.
\newblock Computational and experimental approaches to visual aesthetics.
\newblock \emph{Frontiers in computational neuroscience}, 11\penalty0
  (1):\penalty0 102--119, 2017.

\bibitem[Chang et~al.(2017)Chang, Lu, and Chen]{Chang2017AestheticCG}
K.~Chang, KH. Lu, and CS. Chen.
\newblock Aesthetic critiques generation for photos.
\newblock In \emph{ICCV}, 2017.

\bibitem[Chen et~al.(2020)Chen, Zhang, Zhou, Lei, Xu, Zheng, and
  Fan]{Chen2020AdaptiveFD}
Q.~Chen, W.~Zhang, N.~Zhou, P.~Lei, Y.~Xu, Y.~Zheng, and J.~Fan.
\newblock Adaptive fractional dilated convolution network for image aesthetics
  assessment.
\newblock In \emph{CVPR}, 2020.

\bibitem[Chen et~al.(2017)Chen, Klopp, Sun, Chien, and Ma]{Chen2017LearningTC}
Y.~Chen, J.~Klopp, M.~Sun, S.~Chien, and K.~Ma.
\newblock Learning to compose with professional photographs on the web.
\newblock In \emph{ACM-Multimedia}, 2017.

\bibitem[Cornia et~al.(2018)Cornia, Baraldi, Serra, and
  Cucchiara]{cornia2018predicting}
M.~Cornia, L.~Baraldi, G.~Serra, and R.~Cucchiara.
\newblock {Predicting human eye fixations via an LSTM-based saliency attentive
  model}.
\newblock \emph{IEEE Transactions on Image Processing}, 27\penalty0
  (10):\penalty0 5142--5154, 2018.

\bibitem[Datta et~al.(2006)Datta, Joshi, Li, and Wang]{Datta2006StudyingAI}
R.~Datta, D.~Joshi, J.~Li, and J.~Wang.
\newblock Studying aesthetics in photographic images using a computational
  approach.
\newblock In \emph{ECCV}, 2006.

\bibitem[{Deng} et~al.(2009){Deng}, {Dong}, {Socher}, {Li}, {Kai Li}, and {Li
  Fei-Fei}]{deng2009imagenet}
J.~{Deng}, W.~{Dong}, R.~{Socher}, L.~{Li}, {Kai Li}, and {Li Fei-Fei}.
\newblock Imagenet: A large-scale hierarchical image database.
\newblock In \emph{CVPR}, 2009.

\bibitem[Deng et~al.(2017)Deng, Loy, and Tang]{Deng2017ImageAA}
Y.~Deng, C.C. Loy, and X.~Tang.
\newblock Image aesthetic assessment: An experimental survey.
\newblock \emph{IEEE Signal Processing Magazine}, 34\penalty0 (4):\penalty0
  80--106, 2017.

\bibitem[Dhar et~al.(2011)Dhar, Ordonez, and Berg]{Dhar2011HighLD}
S.~Dhar, V.~Ordonez, and T.~Berg.
\newblock High level describable attributes for predicting aesthetics and
  interestingness.
\newblock In \emph{CVPR}, 2011.

\bibitem[Fang et~al.(2020)Fang, Zhu, Zeng, Ma, and Wang]{fang2020perceptual}
Yuming Fang, Hanwei Zhu, Yan Zeng, Kede Ma, and Zhou Wang.
\newblock Perceptual quality assessment of smartphone photography.
\newblock In \emph{CVPR}, 2020.

\bibitem[Freeman(2007)]{Freeman2007ThePE}
M.~Freeman.
\newblock \emph{The photographer's eye: Composition and design for better
  digital photos}.
\newblock CRC Press, 2007.

\bibitem[He et~al.(2015)He, Zhang, Ren, and Sun]{He2015SpatialPP}
K.~He, X.~Zhang, S.~Ren, and J.~Sun.
\newblock Spatial pyramid pooling in deep convolutional networks for visual
  recognition.
\newblock \emph{IEEE Transactions on Pattern Analysis and Machine
  Intelligence}, 37\penalty0 (9):\penalty0 1904--1916, 2015.

\bibitem[He et~al.(2016)He, Zhang, Ren, and Sun]{he2016deep}
K.~He, X.~Zhang, S.~Ren, and J.~Sun.
\newblock Deep residual learning for image recognition.
\newblock In \emph{CVPR}, 2016.

\bibitem[Hou et~al.(2016)Hou, Yu, and Samaras]{Hou2016SquaredEM}
L.~Hou, C.P. Yu, and D.~Samaras.
\newblock Squared earth mover's distance-based loss for training deep neural
  networks.
\newblock \emph{ArXiv}, abs/1611.05916, 2016.

\bibitem[Hou et~al.(2017)Hou, Cheng, Hu, Borji, Tu, and Torr]{hou2017deeply}
Q.~Hou, MM. Cheng, X.~Hu, A.~Borji, Z.~Tu, and P.~Torr.
\newblock Deeply supervised salient object detection with short connections.
\newblock In \emph{CVPR}, 2017.

\bibitem[Hou and Zhang(2007)]{Hou2007SaliencyDA}
X.~Hou and L.~Zhang.
\newblock Saliency detection: A spectral residual approach.
\newblock In \emph{CVPR}, 2007.

\bibitem[Jahanian et~al.(2015)Jahanian, Vishwanathan, and
  Allebach]{jahanian2015learning}
A.~Jahanian, S.~Vishwanathan, and J.~Allebach.
\newblock Learning visual balance from large-scale datasets of aesthetically
  highly rated images.
\newblock In \emph{Human Vision and Electronic Imaging XX}, 2015.

\bibitem[Jin et~al.(2019)Jin, Wu, Zhao, Li, Zhang, Ge, Zou, Zhou, and
  Zhou]{Jin2019AestheticAA}
X.~Jin, L.~Wu, G.~Zhao, X.~Li, X.~Zhang, S.~Ge, D.~Zou, B.~Zhou, and X.~Zhou.
\newblock Aesthetic attributes assessment of images.
\newblock In \emph{ACM-Multimedia}, 2019.

\bibitem[Joshi et~al.(2011)Joshi, Datta, Fedorovskaya, Luong, Wang, Li, and
  Luo]{joshi2011aesthetics}
Dhiraj Joshi, Ritendra Datta, Elena Fedorovskaya, Quang-Tuan Luong, James~Z
  Wang, Jia Li, and Jiebo Luo.
\newblock Aesthetics and emotions in images.
\newblock \emph{IEEE Signal Processing Magazine}, 28\penalty0 (5):\penalty0
  94--115, 2011.

\bibitem[Ko et~al.(2018)Ko, Lee, and Kim]{ko2018pac}
Keunsoo Ko, Jun-Tae Lee, and Chang-Su Kim.
\newblock {PAC-Net}: Pairwise aesthetic comparison network for image aesthetic
  assessment.
\newblock In \emph{ICIP}, 2018.

\bibitem[Kong et~al.(2016)Kong, Shen, Lin, Mech, and Fowlkes]{Kong2016PhotoAR}
S.~Kong, X.~Shen, Z.~Lin, R.~Mech, and C.~Fowlkes.
\newblock Photo aesthetics ranking network with attributes and content
  adaptation.
\newblock In \emph{ECCV}, 2016.

\bibitem[Lee et~al.(2017)Lee, Kim, Lee, and Kim]{Lee2017SemanticLD}
J.T. Lee, H.~Kim, C.~Lee, and C.~Kim.
\newblock Semantic line detection and its applications.
\newblock In \emph{ICCV}, 2017.

\bibitem[Lee et~al.(2018)Lee, Kim, Lee, and Kim]{Lee2018PhotographicCC}
J.T. Lee, H.~Kim, C.~Lee, and C.~Kim.
\newblock Photographic composition classification and dominant geometric
  element detection for outdoor scenes.
\newblock \emph{Journal of Visual Communication and Image Representation},
  55\penalty0 (1):\penalty0 91--105, 2018.

\bibitem[Li et~al.(2010)Li, Gallagher, Loui, and Chen]{Li2010AestheticQA}
C.~Li, A.~Gallagher, A.~Loui, and T.~Chen.
\newblock Aesthetic quality assessment of consumer photos with faces.
\newblock In \emph{ICIP}, 2010.

\bibitem[Li et~al.(2020{\natexlab{a}})Li, Zhu, Zhao, Ding, and
  Lin]{li2020personality}
Leida Li, Hancheng Zhu, Sicheng Zhao, Guiguang Ding, and Weisi Lin.
\newblock Personality-assisted multi-task learning for generic and personalized
  image aesthetics assessment.
\newblock \emph{IEEE Transactions on Image Processing}, 29\penalty0
  (1):\penalty0 3898--3910, 2020{\natexlab{a}}.

\bibitem[Li et~al.(2020{\natexlab{b}})Li, Li, Zhang, and Zhang]{li2020novel}
Xuewei Li, Xueming Li, Gang Zhang, and Xianlin Zhang.
\newblock A novel feature fusion method for computing image aesthetic quality.
\newblock \emph{IEEE access}, 8:\penalty0 63043--63054, 2020{\natexlab{b}}.

\bibitem[Liu et~al.(2020)Liu, Puri, Kamath, and
  Bhattacharya]{Liu2020CompositionAwareIA}
D.~Liu, R.~Puri, N.~Kamath, and S.~Bhattacharya.
\newblock Composition-aware image aesthetics assessment.
\newblock In \emph{WACV}, 2020.

\bibitem[Liu et~al.(2010)Liu, Chen, Wolf, and Cohen-Or]{liu2010optimizing}
Ligang Liu, Renjie Chen, Lior Wolf, and Daniel Cohen-Or.
\newblock Optimizing photo composition.
\newblock In \emph{Computer Graphics Forum}, 2010.

\bibitem[Lok et~al.(2004)Lok, Feiner, and Ngai]{lok2004evaluation}
S.~Lok, S.~Feiner, and G.~Ngai.
\newblock Evaluation of visual balance for automated layout.
\newblock In \emph{Proceedings of the 9th international conference on
  Intelligent user interfaces}, 2004.

\bibitem[Ma et~al.(2017)Ma, Liu, and Chen]{Ma2017ALampAL}
S.~Ma, J.~Liu, and C.~Chen.
\newblock {A-Lamp}: Adaptive layout-aware multi-patch deep convolutional neural
  network for photo aesthetic assessment.
\newblock In \emph{CVPR}, 2017.

\bibitem[Mai et~al.(2016)Mai, Jin, and Liu]{Mai2016CompositionPreservingDP}
L.~Mai, H.~Jin, and F.~Liu.
\newblock Composition-preserving deep photo aesthetics assessment.
\newblock In \emph{CVPR}, 2016.

\bibitem[Marchesotti et~al.(2011)Marchesotti, Perronnin, Larlus, and
  Csurka]{Marchesotti2011AssessingTA}
L.~Marchesotti, F.~Perronnin, D.~Larlus, and G.~Csurka.
\newblock Assessing the aesthetic quality of photographs using generic image
  descriptors.
\newblock In \emph{ICCV}, 2011.

\bibitem[Martinez and Block(1995)]{Martnez1988VisualFA}
B.~Martinez and J.~Block.
\newblock \emph{Visual forces: an introduction to design}.
\newblock Pearson College Division, 1995.

\bibitem[Murray et~al.(2012)Murray, Marchesotti, and
  Perronnin]{Murray2012AVAAL}
N.~Murray, L.~Marchesotti, and F.~Perronnin.
\newblock {AVA}: A large-scale database for aesthetic visual analysis.
\newblock In \emph{CVPR}, 2012.

\bibitem[Obrador et~al.(2010)Obrador, Schmidt-Hackenberg, and
  Oliver]{obrador2010role}
P.~Obrador, L.~Schmidt-Hackenberg, and N.~Oliver.
\newblock The role of image composition in image aesthetics.
\newblock In \emph{ICIP}, 2010.

\bibitem[Perronnin and Dance(2007)]{Perronnin2007FisherKO}
F.~Perronnin and C.~Dance.
\newblock Fisher kernels on visual vocabularies for image categorization.
\newblock In \emph{CVPR}, 2007.

\bibitem[Pr{\"a}kel(2010)]{Prakel2010TheFO}
D.~Pr{\"a}kel.
\newblock \emph{The fundamentals of creative photography}.
\newblock Bloomsbury Publishing, 2010.

\bibitem[Rawat and Kankanhalli(2015)]{rawat2015context}
Yogesh~Singh Rawat and Mohan~S Kankanhalli.
\newblock Context-aware photography learning for smart mobile devices.
\newblock \emph{ACM Transactions on Multimedia Computing, Communications, and
  Applications}, 12\penalty0 (1):\penalty0 1--24, 2015.

\bibitem[Rawat and Kankanhalli(2016)]{rawat2016clicksmart}
Yogesh~Singh Rawat and Mohan~S Kankanhalli.
\newblock Clicksmart: A context-aware viewpoint recommendation system for
  mobile photography.
\newblock \emph{IEEE Transactions on Circuits and Systems for Video
  Technology}, 27\penalty0 (1):\penalty0 149--158, 2016.

\bibitem[Rawat et~al.(2017)Rawat, Song, and Kankanhalli]{rawat2017spring}
Yogesh~Singh Rawat, Mingli Song, and Mohan~S Kankanhalli.
\newblock A spring-electric graph model for socialized group photography.
\newblock \emph{IEEE Transactions on Multimedia}, 20\penalty0 (3):\penalty0
  754--766, 2017.

\bibitem[Ren et~al.(2017)Ren, Shen, Lin, Mech, and
  Foran]{Ren2017PersonalizedIA}
J.~Ren, X.~Shen, Z.~Lin, R.~Mech, and D.~Foran.
\newblock Personalized image aesthetics.
\newblock In \emph{ICCV}, 2017.

\bibitem[S.~Bhattacharya and Shah(2011)]{Bhattacharya2011AHA}
R.~Sukthankar S.~Bhattacharya and M.~Shah.
\newblock A holistic approach to aesthetic enhancement of photographs.
\newblock \emph{ACM Transactions on Multimedia Computing, Communications, and
  Applications}, 7\penalty0 (1):\penalty0 1--21, 2011.

\bibitem[Savakis et~al.(2000)Savakis, Etz, and Loui]{savakis2000evaluation}
A.~Savakis, S.~Etz, and A.~Loui.
\newblock Evaluation of image appeal in consumer photography.
\newblock In \emph{Human vision and electronic imaging V}, 2000.

\bibitem[Schwarz et~al.(2018)Schwarz, Wieschollek, and Lensch]{schwarz2018will}
Katharina Schwarz, Patrick Wieschollek, and Hendrik~PA Lensch.
\newblock Will people like your image? learning the aesthetic space.
\newblock In \emph{WACV}, 2018.

\bibitem[Su et~al.(2011)Su, Chen, Kao, Hsu, and Chien]{Su2011ScenicPQ}
H.~Su, T.~Chen, C.~Kao, W.~Hsu, and S.~Chien.
\newblock Scenic photo quality assessment with bag of aesthetics-preserving
  features.
\newblock In \emph{ACM-Multimedia}, 2011.

\bibitem[Su et~al.(2021)Su, Vemulapalli, Weiss, Chu, Mansfield, Shapira, and
  Pitts]{su2021camera}
Yu-Chuan Su, Raviteja Vemulapalli, Ben Weiss, Chun-Te Chu, Philip~Andrew
  Mansfield, Lior Shapira, and Colvin Pitts.
\newblock Camera view adjustment prediction for improving image composition.
\newblock \emph{arXiv preprint arXiv:2104.07608}, 2021.

\bibitem[Talebi and Milanfar(2018)]{Talebi2018NIMANI}
H.~Talebi and P.~Milanfar.
\newblock {NIMA}: Neural image assessment.
\newblock \emph{IEEE Transactions on Image Processing}, 27\penalty0
  (8):\penalty0 3998--4011, 2018.

\bibitem[Tang et~al.(2013)Tang, Luo, and Wang]{Tang2013ContentBasedPQ}
X.~Tang, W.~Luo, and X.~Wang.
\newblock Content-based photo quality assessment.
\newblock \emph{IEEE Transactions on Image Processing}, 15\penalty0
  (8):\penalty0 1930--1943, 2013.

\bibitem[Th{\"o}mmes and H{\"u}bner(2018)]{thommes2018instagram}
K.~Th{\"o}mmes and R.~H{\"u}bner.
\newblock Instagram likes for architectural photos can be predicted by
  quantitative balance measures and curvature.
\newblock \emph{Frontiers in psychology}, 9\penalty0 (1):\penalty0 1050--1067,
  2018.

\bibitem[Tu et~al.(2020)Tu, Niu, Zhao, Cheng, and Zhang]{tu2020image}
Yi~Tu, Li~Niu, Weijie Zhao, Dawei Cheng, and Liqing Zhang.
\newblock Image cropping with composition and saliency aware aesthetic score
  map.
\newblock In \emph{AAAI}, 2020.

\bibitem[Wang and Deng(2019)]{Wang2019ModelingHP}
W.~Wang and R.~Deng.
\newblock Modeling human perception for image aesthetic assessment.
\newblock In \emph{ICIP}, 2019.

\bibitem[Wang et~al.(2019)Wang, Yang, Zhang, and Zhang]{Wang2019NeuralAI}
W.~Wang, S.~Yang, W.~Zhang, and J.~Zhang.
\newblock Neural aesthetic image reviewer.
\newblock \emph{IET Computer Vision}, 13\penalty0 (8):\penalty0 749--758, 2019.

\bibitem[Wu et~al.(2017)Wu, Pan, Tsai, Kuo, and Hu]{wu2017high}
Min-Tzu Wu, Tse-Yu Pan, Wan-Lun Tsai, Hsu-Chan Kuo, and Min-Chun Hu.
\newblock High-level semantic photographic composition analysis and
  understanding with deep neural networks.
\newblock In \emph{ICMEW}, 2017.

\bibitem[Wu et~al.(2010)Wu, Bauckhage, and Thurau]{wu2010good}
Yaowen Wu, Christian Bauckhage, and Christian Thurau.
\newblock The good, the bad, and the ugly: Predicting aesthetic image labels.
\newblock In \emph{ICPR}, 2010.

\bibitem[Yoo et~al.(2015)Yoo, Park, Lee, and So~Kweon]{yoo2015multi}
Donggeun Yoo, Sunggyun Park, Joon-Young Lee, and In~So~Kweon.
\newblock Multi-scale pyramid pooling for deep convolutional representation.
\newblock In \emph{CVPRW}, 2015.

\bibitem[Yu et~al.(2018)Yu, Shen, Lin, Mech, and Barnes]{Yu2018LearningTD}
N.~Yu, X.~Shen, L.~Lin, R.~Mech, and C.~Barnes.
\newblock Learning to detect multiple photographic defects.
\newblock In \emph{WACV}, 2018.

\bibitem[Zhang et~al.(2014)Zhang, Gao, Zimmermann, Tian, and
  Li]{Zhang2014FusionOM}
L.~Zhang, Y.~Gao, R.~Zimmermann, Q.~Tian, and X.~Li.
\newblock Fusion of multichannel local and global structural cues for photo
  aesthetics evaluation.
\newblock \emph{IEEE Transactions on Image Processing}, 23\penalty0
  (3):\penalty0 1419--1429, 2014.

\bibitem[Zhao and Wu(2019)]{Zhao_2019_CVPR}
T.~Zhao and X.~Wu.
\newblock Pyramid feature attention network for saliency detection.
\newblock In \emph{CVPR}, 2019.

\bibitem[Zhou et~al.(2016)Zhou, Lu, Zhang, and Wang]{Zhou2016JointIA}
Y.~Zhou, X.~Lu, J.~Zhang, and J.Z. Wang.
\newblock Joint image and text representation for aesthetics analysis.
\newblock In \emph{ACM-Multimedia}, 2016.

\bibitem[Zhou et~al.(2015)Zhou, He, Li, and Wang]{Zhou2015ModelingPE}
Z.~Zhou, S.~He, J.~Li, and J.Z. Wang.
\newblock Modeling perspective effects in photographic composition.
\newblock In \emph{ACM-Multimedia}, 2015.

\end{thebibliography}


\begin{thebibliography}{22}
\providecommand{\natexlab}[1]{#1}
\providecommand{\url}[1]{\texttt{#1}}
\expandafter\ifx\csname urlstyle\endcsname\relax
  \providecommand{\doi}[1]{doi: #1}\else
  \providecommand{\doi}{doi: \begingroup \urlstyle{rm}\Url}\fi

\bibitem[Benjamini and Yekutieli(2001)]{benjamini2001control}
Y.~Benjamini and D.~Yekutieli.
\newblock The control of the false discovery rate in multiple testing under
  dependency.
\newblock \emph{The Annals of Statistics}, 29\penalty0 (4):\penalty0
  1165--1188, 2001.

\bibitem[Chang et~al.(2017)Chang, Lu, and Chen]{Chang2017AestheticCG}
K.~Chang, KH. Lu, and CS. Chen.
\newblock Aesthetic critiques generation for photos.
\newblock In \emph{ICCV}, 2017.

\bibitem[Chen et~al.(2020)Chen, Zhang, Zhou, Lei, Xu, Zheng, and
  Fan]{Chen2020AdaptiveFD}
Q.~Chen, W.~Zhang, N.~Zhou, P.~Lei, Y.~Xu, Y.~Zheng, and J.~Fan.
\newblock Adaptive fractional dilated convolution network for image aesthetics
  assessment.
\newblock In \emph{CVPR}, 2020.

\bibitem[{Deng} et~al.(2009){Deng}, {Dong}, {Socher}, {Li}, {Kai Li}, and {Li
  Fei-Fei}]{deng2009imagenet}
J.~{Deng}, W.~{Dong}, R.~{Socher}, L.~{Li}, {Kai Li}, and {Li Fei-Fei}.
\newblock Imagenet: A large-scale hierarchical image database.
\newblock In \emph{CVPR}, 2009.

\bibitem[Freeman(2007)]{Freeman2007ThePE}
M.~Freeman.
\newblock \emph{The photographer's eye: Composition and design for better
  digital photos}.
\newblock CRC Press, 2007.

\bibitem[He et~al.(2016)He, Zhang, Ren, and Sun]{he2016deep}
K.~He, X.~Zhang, S.~Ren, and J.~Sun.
\newblock Deep residual learning for image recognition.
\newblock In \emph{CVPR}, 2016.

\bibitem[Hou et~al.(2016)Hou, Yu, and Samaras]{Hou2016SquaredEM}
L.~Hou, C.P. Yu, and D.~Samaras.
\newblock Squared earth mover's distance-based loss for training deep neural
  networks.
\newblock \emph{ArXiv}, abs/1611.05916, 2016.

\bibitem[Jin et~al.(2019)Jin, Wu, Zhao, Li, Zhang, Ge, Zou, Zhou, and
  Zhou]{Jin2019AestheticAA}
X.~Jin, L.~Wu, G.~Zhao, X.~Li, X.~Zhang, S.~Ge, D.~Zou, B.~Zhou, and X.~Zhou.
\newblock Aesthetic attributes assessment of images.
\newblock In \emph{ACM-Multimedia}, 2019.

\bibitem[King and Zeng(2001)]{King2001LogisticRI}
G.~King and L.~Zeng.
\newblock Logistic regression in rare events data.
\newblock \emph{Political Analysis}, 9\penalty0 (2):\penalty0 137--163, 2001.

\bibitem[Kingma and Ba(2015)]{kingma2015adam}
D.~Kingma and J.~Ba.
\newblock Adam: A method for stochastic optimization.
\newblock In \emph{ICLR}, 2015.

\bibitem[Kong et~al.(2016)Kong, Shen, Lin, Mech, and Fowlkes]{Kong2016PhotoAR}
S.~Kong, X.~Shen, Z.~Lin, R.~Mech, and C.~Fowlkes.
\newblock Photo aesthetics ranking network with attributes and content
  adaptation.
\newblock In \emph{ECCV}, 2016.

\bibitem[Krishna et~al.(2017)Krishna, Zhu, Groth, Johnson, Hata, Kravitz, Chen,
  Kalantidis, Li, Shamma, et~al.]{krishna2017visual}
R.~Krishna, Y.~Zhu, O.~Groth, J.~Johnson, K.~Hata, J.~Kravitz, S.~Chen,
  Y.~Kalantidis, L.~Li, D.~Shamma, et~al.
\newblock Visual {Genome}: Connecting language and vision using crowdsourced
  dense image annotations.
\newblock \emph{International journal of computer vision}, 123\penalty0
  (1):\penalty0 32--73, 2017.

\bibitem[Murray et~al.(2012)Murray, Marchesotti, and
  Perronnin]{Murray2012AVAAL}
N.~Murray, L.~Marchesotti, and F.~Perronnin.
\newblock {AVA}: A large-scale database for aesthetic visual analysis.
\newblock In \emph{CVPR}, 2012.

\bibitem[Obrador et~al.(2010)Obrador, Schmidt-Hackenberg, and
  Oliver]{obrador2010role}
P.~Obrador, L.~Schmidt-Hackenberg, and N.~Oliver.
\newblock The role of image composition in image aesthetics.
\newblock In \emph{ICIP}, 2010.

\bibitem[Paszke et~al.(2019)Paszke, Gross, Massa, Lerer, Bradbury, Chanan,
  Killeen, Lin, Gimelshein, Antiga, et~al.]{paszke2019pytorch}
A.~Paszke, S.~Gross, F.~Massa, A.~Lerer, J.~Bradbury, G.~Chanan, T.~Killeen,
  Z.~Lin, N.~Gimelshein, L.~Antiga, et~al.
\newblock Pytorch: An imperative style, high-performance deep learning library.
\newblock In \emph{NeurIPS}, 2019.

\bibitem[Pr{\"a}kel(2010)]{Prakel2010TheFO}
D.~Pr{\"a}kel.
\newblock \emph{The fundamentals of creative photography}.
\newblock Bloomsbury Publishing, 2010.

\bibitem[Ren et~al.(2017)Ren, Shen, Lin, Mech, and
  Foran]{Ren2017PersonalizedIA}
J.~Ren, X.~Shen, Z.~Lin, R.~Mech, and D.~Foran.
\newblock Personalized image aesthetics.
\newblock In \emph{ICCV}, 2017.

\bibitem[Ren et~al.(2015)Ren, He, Girshick, and Sun]{Ren2015FasterRT}
S.~Ren, K.~He, R.~Girshick, and J.~Sun.
\newblock Faster {R-CNN}: Towards real-time object detection with region
  proposal networks.
\newblock \emph{IEEE Transactions on Pattern Analysis and Machine
  Intelligence}, 39\penalty0 (6):\penalty0 1137--1149, 2015.

\bibitem[Savakis et~al.(2000)Savakis, Etz, and Loui]{savakis2000evaluation}
A.~Savakis, S.~Etz, and A.~Loui.
\newblock Evaluation of image appeal in consumer photography.
\newblock In \emph{Human vision and electronic imaging V}, 2000.

\bibitem[Talebi and Milanfar(2018)]{Talebi2018NIMANI}
H.~Talebi and P.~Milanfar.
\newblock {NIMA}: Neural image assessment.
\newblock \emph{IEEE Transactions on Image Processing}, 27\penalty0
  (8):\penalty0 3998--4011, 2018.

\bibitem[Wang et~al.(2019)Wang, Yang, Zhang, and Zhang]{Wang2019NeuralAI}
W.~Wang, S.~Yang, W.~Zhang, and J.~Zhang.
\newblock Neural aesthetic image reviewer.
\newblock \emph{IET Computer Vision}, 13\penalty0 (8):\penalty0 749--758, 2019.

\bibitem[Zhou et~al.(2016)Zhou, Lu, Zhang, and Wang]{Zhou2016JointIA}
Y.~Zhou, X.~Lu, J.~Zhang, and J.Z. Wang.
\newblock Joint image and text representation for aesthetics analysis.
\newblock In \emph{ACM-Multimedia}, 2016.

\end{thebibliography}
\end{document}


\maketitle
In this document, we provide additional materials to supplement our main submission. We first present more details about constructing our Composition Assessment DataBase (CADB) in Section \ref{sec:dataset}. Then, we describe the detailed consistency analysis of the collected composition scores in Section \ref{sec:consistency_analysis}, which verifies that our composition quality annotations are reliable for scientific research. Next, we use some examples to illustrate the content bias in the CADB dataset in Section \ref{sec:content_bias}. Meanwhile, in Section \ref{sec:weighted_emd_exp}, we describe the proposed weighted EMD loss and study the effect of using weighted EMD loss to mitigate the content bias. Besides, more implementation details of the proposed method are provided in Section \ref{sec:implement_details}. In Section \ref{sec:hyperparameter_exp}, Section \ref{sec:training_size_exp}, Section \ref{sec:backbone}, Section \ref{sec:pattern_exp}, and Section \ref{sec:more_pattern}, experiments on the hyper-parameter, training set size, backbone, each composition pattern, and using more composition patters further prove the effectiveness of our method. The we compare the performance of our method and human raters in Section \ref{sec:human_ratings}. Finally, in Section \ref{sec:additional_results}, we provide additional visualization results on images inside/outside our CADB dataset.

\section{Our CADB Dataset}
\label{sec:dataset}

\subsection{Data Collection}
\label{sec:data_collection}
Recently, many large-scale aesthetic assessment datasets have been created to facilitate research on image aesthetic evaluation, like Aesthetic Visual Analysis database (AVA) \cite{Murray2012AVAAL}, Aesthetics and Attributes DataBase (AADB) \cite{Kong2016PhotoAR}, Photo Critique Captioning Dataset (PCCD) \cite{Chang2017AestheticCG}, AVA-Comments \cite{Zhou2016JointIA}, AVA-Reviewes \cite{Wang2019NeuralAI}, FLICKER-AES \cite{Ren2017PersonalizedIA}, and DPC-Captions \cite{Jin2019AestheticAA}. Therefore, we can build the CADB dataset upon those existing datasets. \textcolor[rgb]{0,0,0}{Table \ref{table:dataset} provides a summary comparison of CADB to other related aesthetic datasets.} Here we select real photos to construct our dataset, because we target at the real-world application of composition assessment. To the best of our knowledge, among them, only the images in AADB and PCCD datasets are all real photos, while the images in other datasets (\emph{e.g.}, AVA dataset) may be heavily edited or synthetic. Besides, PCCD dataset contains 4,235 images downloaded from a professional photo critique website, most of which are taken by professional photographers and have relatively high composition quality. Differently, AADB dataset provides 10,000 images and contains a much more unbiased distribution of professional photos and amateurish photos. So we choose to construct our CADB dataset based on the AADB dataset.

\begin{table}
    \begin{center}
    \setlength{\tabcolsep}{1mm}{
        \begin{tabular}{|l|c|c|c|c|}
            \hline
            Dataset                             & Images  & All Real Photo   & Composition Score & Raters \\ \hline \hline
            AVA \cite{Murray2012AVAAL}          & 255,530 & N                & N             & -         \\
            PCCD \cite{Chang2017AestheticCG}    & 4,235   & Y                & Y             & 1         \\
            AADB \cite{Kong2016PhotoAR}         & 10,000  & Y                & N             & -         \\ \hline
            CADB(Ours)                          & 9,497   & Y                & Y             & 5         \\
            \hline
        \end{tabular}}
    \end{center}
    \caption{Comparison with existing aesthetic assessment datasets. Our CADB dataset is built upon the AADB dataset, taking into account its all real-world images, and more balanced distribution of professional and amateurish photos \cite{Kong2016PhotoAR}.}
    \label{table:dataset}
    \vspace{-6mm}
\end{table}
\vspace{-3mm}

\subsection{Guidelines for Image Composition Evaluation}
\label{sec:guidelines}
In this section, we show the annotation guidelines for evaluating the quality of image composition, aiming to make the annotation consistent across five individual raters who specialize in fine art. 
1) We provide a composition rating scale from 1 to 5, where a larger score indicates better composition.
2) To help raters quickly learn the rules of image composition rating, we provide them with 100 images with high composition quality and 100 images with low composition quality selected from PCCD dataset \cite{Chang2017AestheticCG} to serve as examples. 
3) When assessing image composition quality, the photographic rules that should be considered are including but not limited to: \emph{rule of thirds}, \emph{centred composition}, \emph{symmetry}, \emph{repetition}, \emph{shallow depth of field}, \emph{diagonals}, \emph{triangles}, \emph{golden ratio}, \emph{frame within the frame}, \emph{leading lines}, \emph{fill the frame}, \emph{isolate the subject},  \emph{vanishing point}, \emph{juxtaposition}, \emph{balancing elements}, and \emph{object emphasis}. In addition, we draw a 3$\times$3 dotted grid on each image as auxiliary lines that divide the image into nine equal rectangles, which is displayed for the raters together with the original image. 
4) The raters are requested to complete the composition rating independently, and the rating procedure for each single image should not be shorter than 20 seconds. 

Besides, the same five raters annotate all images to mitigate the inconsistency across different raters. Following \cite{Murray2012AVAAL, Kong2016PhotoAR}, we average the composition scores of the five raters as the ground-truth composition mean score for each image. 

\begin{figure*}[tbp]
\begin{center}
  \includegraphics[width=1\linewidth]{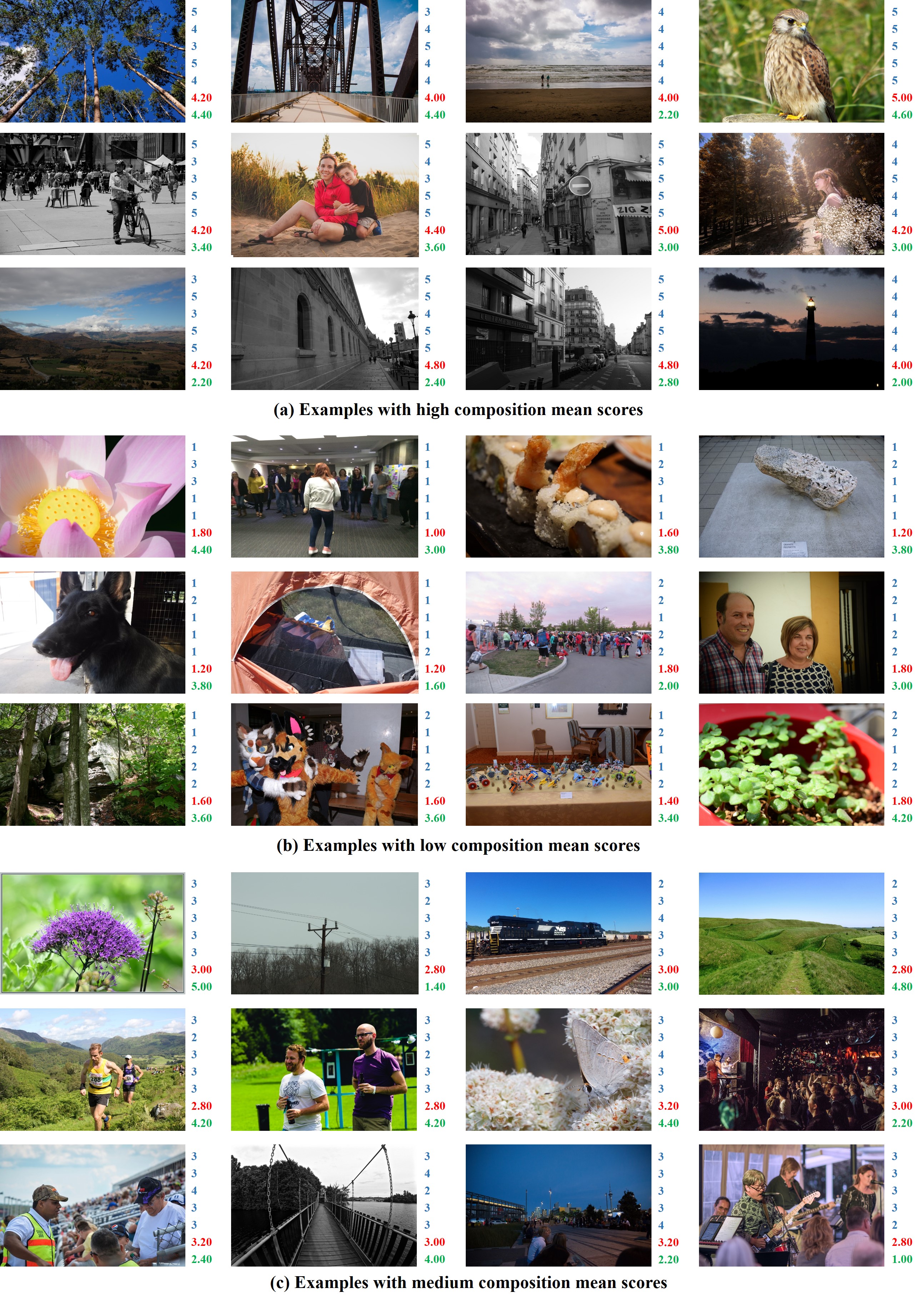}
\end{center}
  \vspace{-5mm}
  \caption{Some example images in our CADB dataset with high/low/medium composition mean scores. We show five composition scores ranging from 1 to 5 provided by five raters in blue and the calculated composition mean score in red. We also show the aesthetic scores annotated by AADB dataset \cite{Kong2016PhotoAR} on a scale from 1 to 5 in green.}
\label{fig:annotation_example}
\end{figure*}
\vspace{-3mm}

\subsection{Annotation Examples and Statistics}
\label{sec:example_stats}
\begin{wrapfigure}{R}{0.5\linewidth}
  \vspace{-9mm}
  \begin{center}
    \includegraphics[width=1\linewidth]{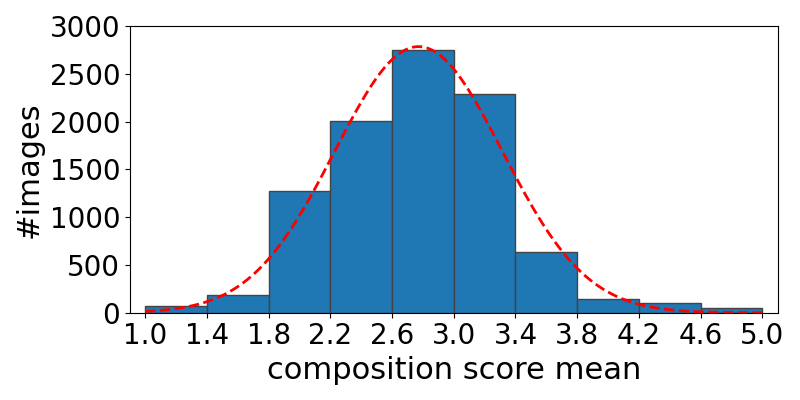}
  \end{center}
  \vspace{-6mm}
  \caption{The distribution of composition mean score in our CADB dataset. The dashed line indicates the fitted Gaussian distribution.}
  \vspace{-2mm}
  \label{fig:score_distribution}
\end{wrapfigure}

In Figure \ref{fig:annotation_example}, we present some examples in our CADB dataset with five composition scores and composition mean score that is obtained by averaging those composition scores for each image. For better visualization, we divide these examples into three groups according to the composition mean score: images with high, low, medium scores. From Figure \ref{fig:annotation_example}, we can roughly verify the validness of the composition annotations.

In Figure \ref{fig:annotation_example}, we also present the aesthetic scores annotated by AADB dataset \cite{Kong2016PhotoAR} , which is also rated by multiple individual human raters on a scale from 1 to 5 for the overall aesthetic quality, a larger score indicates higher aesthetic quality. We can see that, for some images, a high (\emph{resp.}, low) composition score does not mean a high (\emph{resp.}, low) aesthetic score. This is because that the composition assessment focuses on analyzing the placement of visual elements in the image, while aesthetic evaluation quantifies the aesthetic quality of the image in a comprehensive manner by taking not only image composition but also other visual factors (\emph{e.g.}, interesting content, good lighting, color harmony, vivid color, motion blur, and shallow depth of field) into consideration. The essential difference between the above tasks further sheds light on the significance of specially developing methods for evaluating overall composition quality.

Furthermore, we calculate the distribution of composition mean score in Figure \ref{fig:score_distribution}, where indicates that the scores are well fit by Gaussian distribution similar to the observation in AADB dataset \cite{Kong2016PhotoAR} and AVA dataset \cite{Murray2012AVAAL}. As shown in Figure \ref{fig:score_distribution}, the average and variance of composition mean score is 2.70 and 0.35, respectively.

\section{Consistency Analysis of Annotations}
\label{sec:consistency_analysis}
\begin{figure*}[tbp]
\begin{center}
  \includegraphics[width=1.0\linewidth]{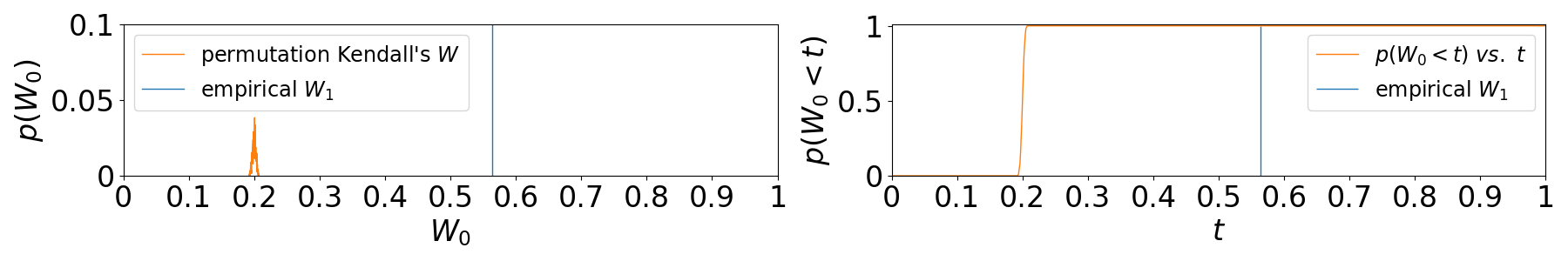}
\end{center}
  \vspace{-4mm}
  \caption{Permutation test on Kendall's $W$. Left: $p(W)$ \emph{vs.} $W$. Right: $p(W < t)$ \emph{vs.} $t$.}
  \vspace{-4mm}
\label{fig:kindall_pw}
\end{figure*}
Considering the subjective nature of human aesthetic activity \cite{savakis2000evaluation,Prakel2010TheFO,Freeman2007ThePE}, we carry out consistency analysis on the composition scores provided by multiple raters to verify that our CADB dataset is qualified for scientific evaluation. 
Following \cite{Kong2016PhotoAR}, we employ both Kendall's concordance coefficient (also known as Kendall's $W$) and Spearman's rank correlation coefficient (also known as Spearman's $\rho$) in the experiments. Kendall's $W$ indicates the agreement among multiple raters and accounts for tied ranks, the value of which varies from 0 (no agreement) to 1 (complete agreement). Spearman's $\rho$ is computed between the predicted and ground-truth composition score distribution to measure their closeness. 

Since five raters annotated a collection of 10,200 images (including 240 sanity check images), we calculate an average Kendall's $W$ of $0.5734$ over all images, which demonstrates significant consistency among different raters. Then, following \cite{Kong2016PhotoAR}, we conduct a permutation test over global Kendall's $W$ to obtain the distribution of $W$ under the null hypothesis, the curves of which $p(W)\ \emph{vs.}\ W$ and $p(W < t)\ \emph{vs.}\ t$ are illustrated in Figure \ref{fig:kindall_pw}. We can observe that the empirical Kendall's $W$ on our CADB dataset is statistically significant from both curves.

Then, similar to \cite{Kong2016PhotoAR}, we investigate the consistency of composition scores at batch level and randomly split all annotated samples into multiple batches with each batch containing 100 images. For each batch, we calculate Kendall's $W$ to evaluate the consistency of annotations provided by different raters and confirm its statistical significance by using Benjamini-Hochberg procedure \cite{benjamini2001control} controlling the false discovery rate (FDR) for multiple comparisons. At FDR level $Q=0.05$, $100.0\%$ batches have significant agreement, which means that all batches of our annotations have consistent composition scores, and our dataset is qualified for scientific evaluation. 

Moreover, we also adopt Spearman's $\rho$ to measure the rank correlation between the composition scores of pairwise raters and test its statistical significance at batch level via Benjamini-Hochberg procedure. The $p$-value for each batch is computed by averaging the pairwise $p$-values in the current batch following \cite{Kong2016PhotoAR}. At FDR level $Q=0.05$, $98.04\%$ batches have significant agreement, which further confirms the reliability of the composition quality annotations in our CADB dataset. 
\vspace{-4mm}
\section{Content Bias}
\label{sec:content_bias}

\begin{figure*}[tbp]
\begin{center}
  \includegraphics[width=1\linewidth]{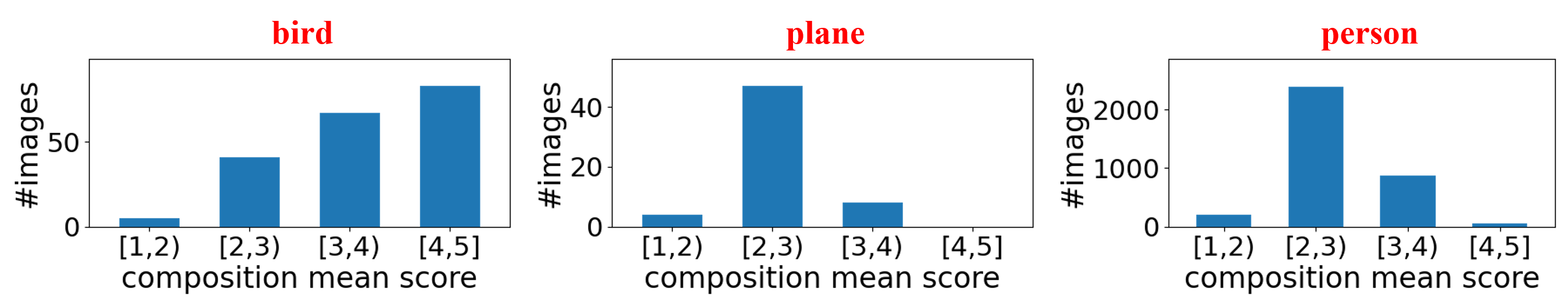}
\end{center}
  \vspace{-4mm}
  \caption{Illustration of biased categories in the CADB dataset. For each category (colored red), given images containing this object category, we count the occurrence of composition mean score in four score bins.}
  \vspace{-4mm}
\label{fig:content_bias}
\end{figure*}

In Section {\color{red}3} of the main text, we briefly mention the content bias issue in our CADB dataset. Here we provide a detailed description of this concept. Intuitively, photos of any object category have chances to be of high or low composition quality, which means that composition mean score $\bar{y}$ should be approximately evenly distributed within each category. However, in our CADB dataset, as illustrated in Figure \ref{fig:content_bias}, we observe that there are some categories whose score distributions are concentrated in a very narrow interval, and we refer to these categories as biased categories. For example, as shown in in Figure \ref{fig:content_bias}, most bird photos are rated with high scores, probably because bird photos are more likely to be taken by professional photographers rather than amateurs. In this case, the network may find a shortcut to simply rate images based on their contents, which is dubbed as content bias in this paper.

To identify the biased categories, we first leverage Faster R-CNN \cite{Ren2015FasterRT} trained on Visual Genome \cite{krishna2017visual} to detect objects for all images. We divide the range of composition mean score $\bar{y}$ into $M$ bins ($M=4$) with the bin size equal to $1$ (\emph{e.g.}, $[1,2)$). For the images containing each object category, we count the occurrence of $\bar{y}$ in each bin and derive the score distribution over $M$ bins. To measure the degree of bias, we compute the entropy of the score distribution for all categories. The category with an entropy below $0.1$ is treated as a highly biased category whose associated images will be removed from our dataset. After this step, there are 9,497 images left. Then, we calculate the ratio of the maximum occurrence to the minimum non-zero occurrence in the $M$ bins as $r_c$ for the $c$-th category. A category is defined as an unbiased category if $r_c \leq 1.5$ and otherwise a biased category. Furthermore, an image is defined as an unbiased image if its involving categories are all unbiased categories and otherwise a biased image.

For the test images in real-world applications, photos of any object category have chances to be of high or low composition quality. For better evaluation, we select 950 unbiased images to form the test set, which is closer to the test set in practice, and use the remaining 8,547 images (including both unbiased and biased ones) for training.

\section{Weighted EMD Loss}
\label{sec:weighted_emd_exp}

In our CADB dataset, each image is rated by five raters, so both composition mean score and composition score distribution can be computed. Considering the intrinsic orderliness of our composition rating scale (see Section~\ref{sec:guidelines}), we train our model to predict composition score distribution and adopt the normalized EMD loss \cite{Hou2016SquaredEM}, which has been widely used in aesthetic assessment \cite{Talebi2018NIMANI, Chen2020AdaptiveFD}. We assume that the ground-truth and predicted composition score distribution are $\mathbf{y}$ and $\mathbf{\hat{y}}$, respectively. Then, the normalized EMD loss can be calculated by
\begin{equation}
    \mathcal{L}_{EMD}(\mathbf{y}, \hat{\mathbf{y}}) = \left( \frac{1}{S} \sum^{S}_{s=1} \left | \mathrm{CDF}_{\mathbf{y}}(s) - \mathrm{CDF}_{\mathbf{\hat{y}}}(s) \right | ^{r} \right )^{1/r},
\label{emd}
\end{equation}
where $S=5$ is the scale of composition score in our dataset and $r$ is a hyper-parameter. $\mathrm{CDF}_{\mathbf{y}}(s)=\sum^{s}_{i=1} y_i$ denotes the cumulative distribution function. We set $r=2$ following \cite{Talebi2018NIMANI, Chen2020AdaptiveFD}. The predicted composition mean score can be calculated as the expectation of the score distribution $\sum^{S}_{i=1} i \cdot \hat{y}_i$.
\begin{figure*}[tbp]
\begin{center}
   \includegraphics[width=1.0\linewidth]{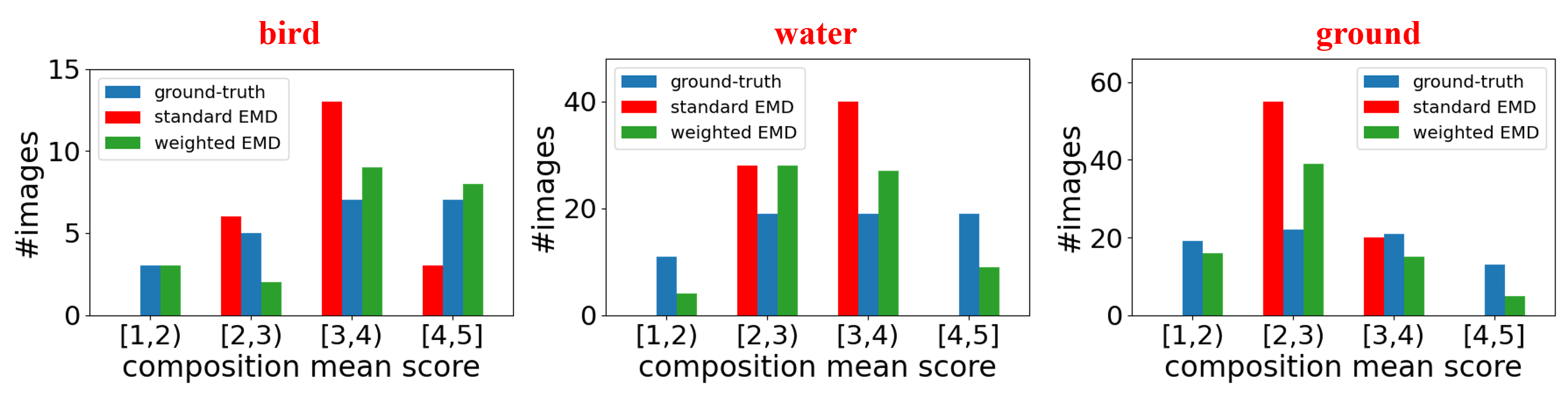}
\end{center}
  \vspace{-4mm}
   \caption{Illustration of using weighted EMD loss to eliminate content bias. For each category (colored red), given images containing this object category, we count three type of occurrences of composition mean scores in four score bins: ground-truth, predicted by the model trained with EMD loss, predicted by the model trained with our weighted EMD loss.}
  \vspace{-2mm}
\label{fig:demonstrate_emd}
\end{figure*}
As discussed in Section~\ref{sec:content_bias}, we observe content bias in our dataset, that is, the images with certain object categories are more likely to have high or low composition scores.
Training on such data, the network may find a shortcut to simply rate images based on their contents, leading to weak generalization ability to real-world photos. 
To eliminate the effect of content bias and prevent the model from learning a shortcut, we propose a strategy that assigns different weights to different samples when calculating EMD Loss. Specifically, as mentioned in Section~\ref{sec:content_bias}, the range of composition mean score $\bar{y}$ is divided into $M$ bins. We use $T_{m,c}$ to present the occurrence that the $c$-th category appears in $m$-th bin and calculate weights for each category via the strategy proposed in \cite{King2001LogisticRI}: $\alpha_{m,c} = \frac{\sum^{M}_{m=1} T_{m,c}}{M \cdot T_{m,c}}$, which is inversely proportional to $T_{m,c}$.
Given an image that contains $C$ object categories and has a $\bar{y}$ falling in the $m$-th bin, we take the minimum weight across all categories as its weight $\beta = \min \{ \alpha_{m,1},  \alpha_{m,2}, \ldots, \alpha_{m,C} \}$. Instead of \emph{minimum}, we have also tried several other options (\emph{e.g.}, \emph{maximum}, \emph{median}, and \emph{mean}), but \emph{minimum} gives the best result. The weight $\beta$ is different for different training samples. We precompute $\beta$ for all training samples beforehand and assign sample-specific weight $\beta$ to EMD loss (\ref{emd}) during training. 

In the ablation study in Section {\color{red} 5.2} of the main text, we have confirmed that using weighted EMD loss can benefit model performance by eliminating content bias. To take a closer look at the advantage of weighted EMD loss in eliminating content bias, for each category, we analyze the distribution of ground-truth/predicted composition mean scores of images containing this object category.

Specifically, we first employ the ResNet18 \cite{he2016deep} backbone trained with EMD loss (\emph{resp.}, weighted EMD loss) to estimate composition mean scores for images in the testing set. Then, for each category, we collect the ground-truth/predicted composition mean scores of images containing this object category. After that, we visualize the distribution of composition mean score for each category in a similar way to Section \ref{sec:content_bias}. As illustrated in Figure \ref{fig:demonstrate_emd}, for each example category, we show three types of composition mean scores: ground-truth, predicted by the model trained with EMD loss, predicted by the model trained with proposed weighted EMD loss. Comparing the results of EMD loss and weighted EMD loss, we can see that training with weighted loss produces a much more unbiased distribution of composition mean score, which also looks closer to the ground-truth distribution. For example, for the results on water images in Figure \ref{fig:demonstrate_emd}, the ground-truth composition mean scores are approximately evenly distributed across four bins, while the composition mean score distribution of using EMD loss is concentrated in the intervals of [2,3) and [3,4). Differently, for the model trained with weighted EMD loss, the predicted composition mean score distribution is relatively balanced on all four bins, from which we can validate the advantages of the proposed weighted EMD loss on eliminating content bias qualitatively. 

\begin{figure}[tbp]
\begin{center}
   \includegraphics[width=1\linewidth]{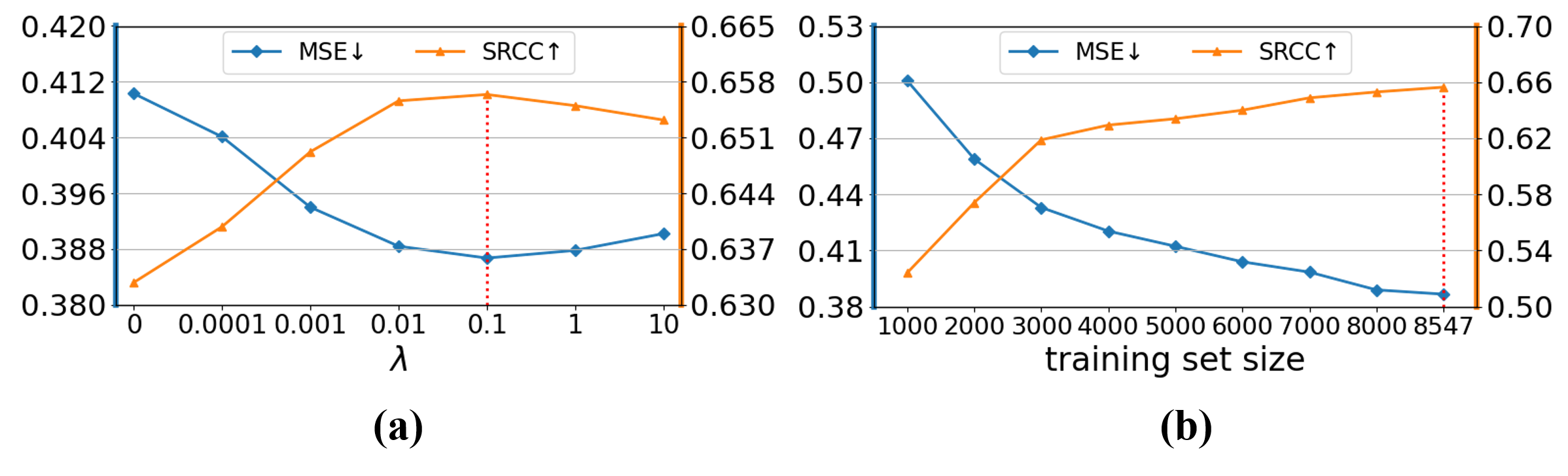}
\end{center}
   \vspace{-8mm}
   \caption{Analysis of hyper-parameters and training set size on our CADB dataset. (a) Performance variation of our model with different hyper-parameter $\lambda$. (b) Performance variation of our model with different training set size. The dashed vertical line denotes the default value used in our paper.}
  \vspace{-2mm}
\label{fig:hyper_and_training_size}
\end{figure}

\begin{table}
    \begin{center}
    \setlength{\tabcolsep}{5.5mm}{
        \begin{tabular}{|l|c|c|c|c|}
            \hline
            Backbone   & MSE$\downarrow$    &EMD$\downarrow$    &SRCC$\uparrow$     &LCC$\uparrow$    \\ \hline \hline
            ResNet-18  & 0.3867             & 0.1798            & 0.6564            & 0.6709          \\
            ResNet-34  & \textbf{0.3776}    & \textbf{0.1794}   & \textbf{0.6736}   & \textbf{0.6808} \\
            ResNet-50  & 0.399              & 0.1819            & 0.6539            & 0.6595          \\
            ResNet-101 & 0.4059             & 0.1824            & 0.6463            & 0.6563          \\
            \hline
        \end{tabular}}
    \end{center}
    \caption{Performance of our method with different backbone networks.}
    \label{table:backbone}
    \vspace{-4mm}
\end{table}

\section{Implementation Details}
\label{sec:implement_details}
We implement our model and conduct all experiments using Pytorch \cite{paszke2019pytorch}. During the training stage, the backbone weights are pretrained on ImageNet \cite{deng2009imagenet} and other layers are randomly initialized. 
We adopt the Adam optimizer \cite{kingma2015adam} and set the batch size as 16. Then, the initial learning rate of the layers in the backbone and the layers in the additional modules (\emph{e.g.}, SAMP, AAFF, and prediction head) are set as $1e^{-6}$ and $1e^{-4}$, respectively. This is because we noticed that using a small learning rate on the backbone results in easier and faster optimization in our experiments. Moreover, the learning rate of all layers is annealed by 0.1 every time the training loss plateaus. To prevent overfitting, a dropout rate of 0.5 is applied on each fully-connected layer of the additional modules, and we set weight decay as $5e^{-5}$ for all layers in our network.

\section{Hyper-parameter Analysis}
\label{sec:hyperparameter_exp}
There is a trade-off parameter $\lambda$ before the attribute loss in Eq.({\color{red}1}) of the main text. We set the hyper-parameter according to cross-validation by splitting 20\% training samples as validation set. We vary $\lambda$ from 0 to 10 and present the results in Figure \ref{fig:hyper_and_training_size}(a), in which we report Mean Squared Error (MSE) and Spearman's Rank Correlation Coefficient (SRCC). Comparing the result without attribute loss ($\lambda=0$) and the result with $\lambda=0.1$, we can see a clear gap between their performance. Therefore, we set $\lambda=0.1$ by default for all experiments. Moreover, the experimental results demonstrate that our method is robust when setting $\lambda$ in the range of [0.01,10].

\begin{figure}[tbp]
\begin{center}
  \includegraphics[width=0.7\linewidth]{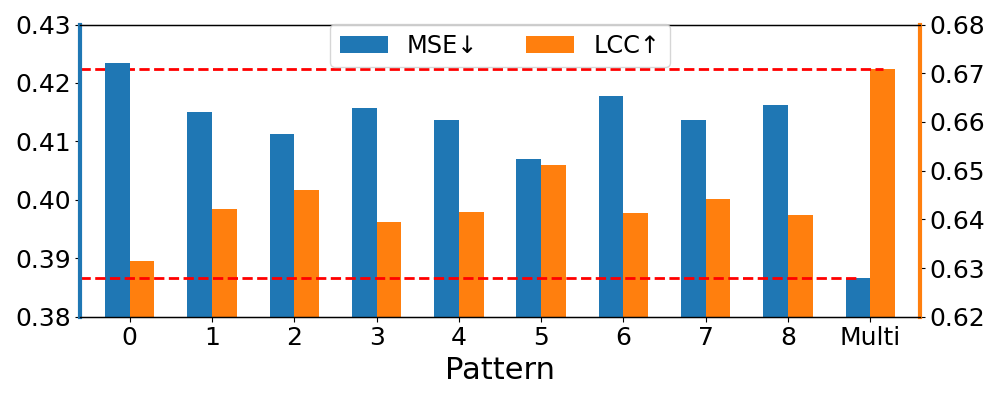}
\end{center}
  \vspace{-6mm}
  \caption{Results of our model using each individual composition pattern (pattern 1$\sim$ 8). Pattern 0 means the simplest pattern with only one partition. \emph{Multi} means using all 8 patterns.}
  \vspace{-6mm}
\label{fig:pattern_analysis}
\end{figure}

\section{Different Training Set Size}
\label{sec:training_size_exp}

As mentioned in Section~{\color{red}3} of the main text, we split the CADB dataset into training (8,547) and test (950) sets. To study the correlation between the test performance of our model and the training set size, we randomly select a certain amount of samples from training set to train the model and evaluate on the same test set. We vary the number of training samples from 1,000 to 8,000 with the step length of 1,000 and report the results in Figure \ref{fig:hyper_and_training_size}(b) using MSE and SRCC. When the training set size increases, the model performance improves significantly, yet the performance growth slows down. When the training size gets larger than 8,000, the performance gain becomes negligible, demonstrating that our model capacity is compatible with the CADB dataset.

\section{Different Backbone Network}
\label{sec:backbone}

\textcolor[rgb]{0,0,0}{We evaluate our method with different backbones on the CADB dataset and report results in Table \ref{table:backbone}. It can be see that our method achieves the best result using ResNet-34 and the performance drop using ResNet-50 or ResNet-101 might be caused by overfitting. Given that ResNet-18 is efficient and can already receive good results, we adopt ResNet-18 as the default backbone in our method.} 

\section{Effectiveness of Composition Pattern}
\label{sec:pattern_exp}

Recall that we design eight composition patterns (see Figure {\color{red}3(a)} of the main text) for composition evaluation from different perspectives.
To study the effectiveness of each pattern, we conduct experiments on our SAMP-Net with only a single pattern in SAMP. Moreover, we compare with the simplest pattern with only one partition (\emph{i.e.}, global pooling), which is referred to as pattern $0$. The experimental results are summarized in Figure \ref{fig:pattern_analysis}, where we report MSE and Linear Correlation Coefficient (LCC). It can be seen that all the models with the designed patterns, including single pattern and multi-pattern, perform better than pattern $0$, which indicates that our designed patterns are meaningful and helpful for composition assessment. Among the results using a single pattern, we find that pattern $5$ performs best in terms of MSE, which might because visually important objects are often placed at the center of images. Furthermore, the multi-pattern model beats all single-pattern models, which again demonstrates the effectiveness of our SAMP module.

\begin{figure*}[tbp]
\begin{center}
   \includegraphics[width=1.0\linewidth]{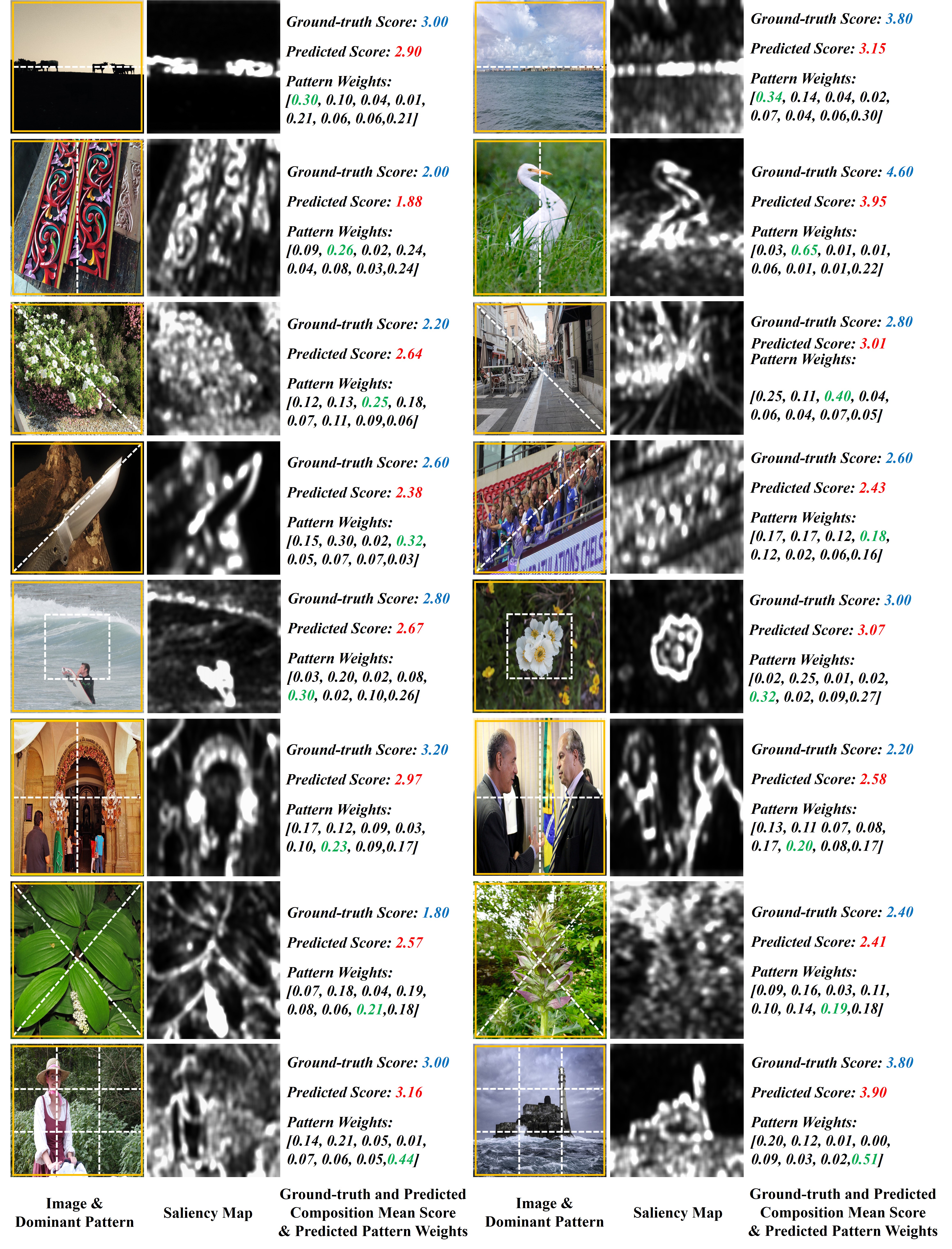}
\end{center}
   \caption{Visualization results of the proposed method on our CADB dataset. We show the estimated pattern weights and the largest weight is colored green. We also show the ground-truth/predicted composition mean score in blue/red. }
\label{fig:more_results}
\end{figure*}

\begin{figure*}[tbp]
\begin{center}
   \includegraphics[width=1.0\linewidth]{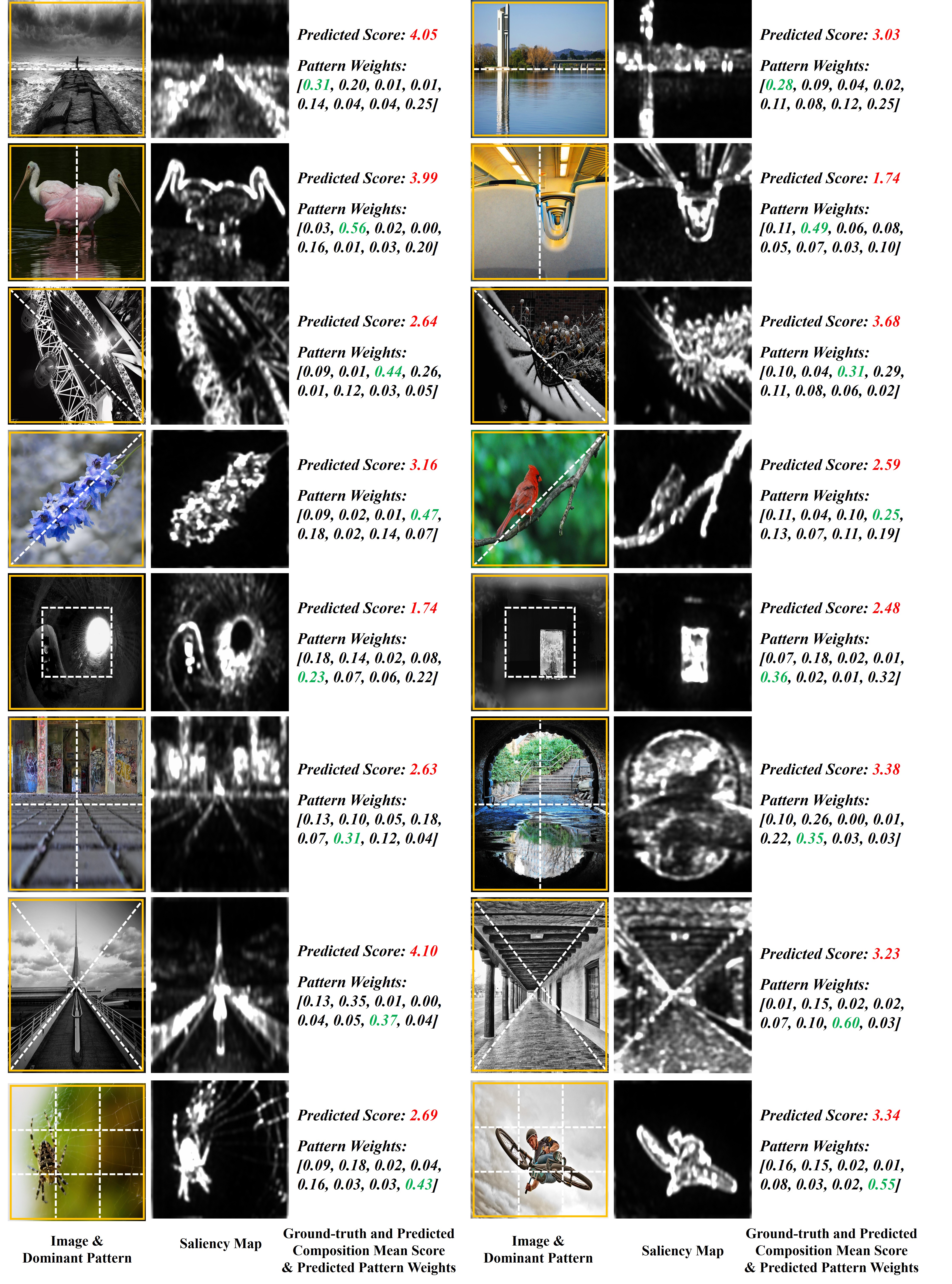}
\end{center}
   \caption{Visualization results of the proposed method on the PCCD dataset \cite{Chang2017AestheticCG}. We show the estimated pattern weights and the largest weight is colored green. We also show the predicted composition mean score in red. }
\label{fig:results_on_pccd}
\end{figure*}

\section{Using More Composition Patterns}
\label{sec:more_pattern}

\begin{table*}
    \begin{center}
    \setlength{\tabcolsep}{2.5mm}{
        \begin{tabular}{|c|c|c|c|c|}
            \hline
            Patterns &   MSE$\downarrow$   &EMD$\downarrow$    &SRCC$\uparrow$     &LCC$\uparrow$ \\
            \hline\hline            
            Patterns 1$\sim$ 8  & \textbf{0.3867}    &\textbf{0.1798}    &\textbf{0.6564}    &\textbf{0.6709} \\
            Patterns 1$\sim$ 11 &     0.3876           & 0.1800            & 0.6558            & 0.6701       \\
            \hline
        \end{tabular}}
    \end{center}
    \caption{Results of using more composition patterns. Based on the existing eight composition patterns, we add three more composition patterns (see Figure \ref{fig:new_patterns}) in our model.}
    \label{table:more_pattern}
    \vspace{-2mm}
\end{table*}

\begin{wrapfigure}{R}{0.3\linewidth}
  \vspace{-7mm}
  \begin{center}
    \includegraphics[width=1\linewidth]{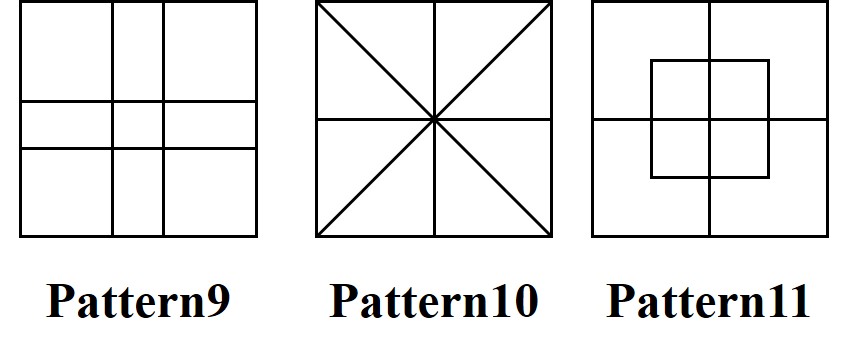}
  \end{center}
  \vspace{-4mm}
  \caption{Three additional composition patterns.}
  \label{fig:new_patterns}
\end{wrapfigure}
\textcolor[rgb]{0,0,0}{Apart from existing eight patterns (see Figure {\color{red}3(a)} of the main text), to evaluate the effect of learning more diverse rules, we design three additional composition patterns in Figure \ref{fig:new_patterns}. Pattern 9 is inspired by gloden ratio \cite{obrador2010role}. Pattern 10 and pattern 11 concern more complex composition patterns.
The results in Table \ref{table:more_pattern} implies that using more composition patterns cannot achieve further improvement. We would like to explore other  more composition patterns in the future.}

\section{Comparison with Human Ratings}
\label{sec:human_ratings}
\begin{table}
    \begin{center}
    \setlength{\tabcolsep}{5.5mm}{
        \begin{tabular}{|c|c|c|c|c|}
            \hline
            Human Rater    & MSE$\downarrow$    &EMD$\downarrow$    &SRCC$\uparrow$     &LCC$\uparrow$    \\ \hline \hline
            1        & 0.1951          & 0.1403          & 0.8925         & 0.8944          \\
            2        & 0.4286          & 0.2089          & 0.7694         & 0.7811          \\
            3        & 0.5345          & 0.2328          & 0.7688         & 0.7705          \\
            4        & \textbf{0.1606} & \textbf{0.1369} & \textbf{0.8990} & \textbf{0.9043} \\
            5        & 0.1814          & 0.1430          & 0.8874         & 0.8934          \\    \hline
            SAMP-Net & 0.3867          & 0.1798          & 0.6564         & 0.6709          \\ 
            \hline
        \end{tabular}}
    \end{center}
    \caption{Comparison with human raters on the CADB dataset.}
    \label{table:human_performance}
    \vspace{-4mm}
\end{table}
\textcolor[rgb]{0,0,0}{
We have shown that the proposed method outperforms existing methods in the Section {\color{red}5.3} of the main text. 
To further analyze the capability of our method, we evaluate the performance of each individual raters by comparing with the ground-truth in the same way. Unlike our model which predicts a composition score distribution, each rater only has one score for each test image, resulting in an one-hot score distribution. 
We summarize the results in Table \ref{table:human_performance}. 
Interestingly, our method can outperform two of five human raters in terms of  MSE and EMD. This may be due to the fact that our model is trained using the ratings of all raters and the prediction is close to the average distribution. However, considering SRCC and LCC, which indicate the ability to correctly rank different images according to their composition quality, we see that there is still a clear gap between our model and human raters.} 

\section{Additional Visualization Results}
\label{sec:additional_results} 

 Our SAMP-Net can facilitate composition assessment by integrating the information from multiple patterns and provide constructive suggestions for improving the composition quality. So we present additional examples in Figure \ref{fig:more_results}, in which we show the input image, its saliency map, its ground-truth/predicted composition mean score, and its pattern weights. We refer to the composition pattern with the largest weight as the dominant pattern of the input image. For each pattern, we present two example images with this pattern as dominant pattern and draw this pattern on the image. 
 
 As discussed in Section {\color{red}5.4} of the main text, the dominant pattern unveils from which perspective the input image is given a high or low score. For example, in Figure \ref{fig:more_results}, in the right column of the second row, the vertical line of pattern 2 is parallel to the bird of the image, which looks more visually assuring to viewers. In the left column of the fourth row, pattern 4 implies that the knife is organised based on the diagonal line in the image. Since such images create a sense of visual balance and stability for viewer, the model estimates a relatively high score for them. On the contrary, in the left column of the second row in Figure \ref{fig:more_results}, the carvings slightly deviate from their symmetrical axis under pattern 2. So the low score implies that maintaining horizontal symmetry may help to improve the composition quality. 
 In the left column of the fifth row, per the relatively low score under pattern 5, the surfer is suggested to be moved towards the center.
 Those examples further validate the utility of our model for providing interpretable composition guidance.
 
 Furthermore, we also test our model on some images outside the CADB dataset to show the generalization ability. Specifically, we test our model on some images collected from PCCD dataset \cite{Chang2017AestheticCG} and show the results in Figure \ref{fig:results_on_pccd}. Although the PCCD dataset contains the overall composition score, they only present one reviewer's composition rating for each image and this reviewer (an anonymous website visitor) may be unprofessional, rendering the composition annotations of PCCD  very noisy. Thus, we only report the composition score estimated by our model in Figure \ref{fig:results_on_pccd}. We can see that our model can reasonably predict composition mean scores.

\bibliography{egbib}